\documentclass[12pt]{article}

\usepackage{tensor_style}
\usepackage{multirow}
\usepackage{amsmath,amsfonts}
\usepackage{algorithmic}
\usepackage{graphicx}
\usepackage{textcomp}
\usepackage{xcolor}
\usepackage{booktabs}
\usepackage[top=1in, bottom=1in, right=1in, left=1in]{geometry}
\usepackage{amsthm}
\usepackage[font={small,bf}]{caption}
\usepackage[font={small,bf}]{subcaption}
\usepackage{mdframed}
\newmdtheoremenv{theo}{Theorem}
\usepackage{resizegather}
\usepackage{float}
\usepackage{hyperref}

\hypersetup{
    colorlinks=true,  %
    linkcolor=red, %
    citecolor=red,   %
    urlcolor=red,      %
    filecolor=red   %
}

\usepackage{setspace}

\usepackage{natbib}

\title{Analyzing Political Text at Scale with Online Tensor LDA}
\author{Sara Kangaslahti,$^{1\ast}$ Danny Ebanks,$^{1\ast}$ Jean Kossaifi,$^{2}$, Anqi Liu,$^{3}$ \\R. Michael Alvarez$^{4}$ and Animashree Anandkumar$^{4}$\\
\\
\normalsize{$^{1}$Harvard University, USA}\\
\normalsize{$^{2}$NVIDIA, USA}\\
\normalsize{$^{3}$Johns Hopkins University, USA}\\
\normalsize{$^{4}$California Institute of Technology, USA}\\
}

\date{\today}

\begin{document}

\maketitle

\begin{abstract}

This paper proposes a topic modeling method that scales linearly to billions of documents.
We make three core contributions: i) we present a topic modeling method, Tensor Latent Dirichlet Allocation (TLDA), that has identifiable and recoverable parameter guarantees and sample complexity guarantees for large data; ii) we show that this method is computationally and memory efficient (achieving speeds over 3-4x those of prior parallelized Latent Dirichlet Allocation (LDA) methods), and that it scales linearly to text datasets with over a billion documents; iii) we provide an open-source, GPU-based implementation, of this method.  This scaling enables previously prohibitive analyses, and we perform two real-world, large-scale new studies of interest to political scientists: we provide the first thorough analysis of the evolution of the \#MeToo movement through the lens of over two years of Twitter conversation and a detailed study of social media conversations about election fraud in the 2020 presidential election.  Thus this method provides social scientists with the ability to study very large corpora at scale and to answer important theoretically-relevant questions about salient issues in near real-time. 

\end{abstract}

\doublespacing

\section{Introduction}
We propose a new method to estimate topic models that is feasible on large scale data that has theoretical accuracy guarantees. Our approach leverages theoretical insights from \cite{anandkumar2013spectral} that show that a spectral decomposition approach to topic models scales to large datasets and possesses desirable theoretical properties, such as provable, accurate recovery of the parameters and large-sample consistency. To achieve scale, we show that by demeaning and batching the data,  our method estimates topic model outputs for large scale documents and recovers the same model as \cite{anandkumar2013spectral}, endowing it with the same theoretical guarantees. This approach has many benefits for political scientists, who have used topic modeling methods to study important questions across the discipline, such as studies using text data to study new questions concerning political behavior \citep{barbera2015tweeting,metzgeretal2016,munger2017} and public opinion \citep{barbera2015birds,barbera2019}.  Text data has been important for new advances in analyzing the evolution of protest movements and social protests  \citep{kannetal2023,larsonetal2019,steinertthrelkeld2017,tillery2019}. New methods for analyzing text data and more accessible data have allowed researchers to explore political communications, agenda setting, and the news media \citep{Barbera_Boydstun_Linn_McMahon_Nagler_2021, gilardi2022}.  

In this paper, we contribute to a rich line of methodological research in political science that has innovated and proposed clever frameworks to meet the needs of applied researchers across a wide variety of domains. From best practices and research design frameworks for how to incorporate text \citep{grimmer2013,grimmeretal2022,HopkinsKing2010}, to approaches to unsupervised methods \citep{denny2018}, political methodologists are guiding the field in how to best approach this high dimensional data. Researchers have also introduced new tools for political scientists, including topic models which incorporate metadata \citep{roberts2014}, computer-assisted techniques in processing texts in both clustering and in comparative settings \citep{grimmer2011,lucas2015}, lexical feature selection  \citep{monroe2017}, and crowd-sourcing approaches to measure sophistication \citep{benoit2019}. For example, in this paper, we study two large-scale political science datasets. First, we study a dataset comprised of tweets generated before and during the \#MeToo movement to better understand the evolution of collective action and protest movements. The Women's March in January 2017 was the largest political protest in American history up until that time \cite{Atlantic2017}. We also study a dataset of tweets generated after the 2020 Presidential election in order to better understand coordination effects, the loser's effect, and how online publics react to electoral defeat. These large, dynamic, and unstructured datasets offer new insights into mass-politics and how it manifests on online discourse in particular. This is especially helpful where political scientists have been limited to surveys which rely on respondent recall and are generally static in nature.

Importantly, researchers are now collecting text datasets that are larger and larger in scale.  For example, there are now numerous studies from different disciplines reporting the use of datasets that contain more than a billion tweets \citep{dimitrov2020,Hannak2021,Sinnenberg2016}. 
However, for large datasets, typical approaches for the estimation of LDA methods are often computationally impractical and memory inefficient. Data discovery and description techniques for these large data can help inform new theoretical frameworks, establish critical empirical facts, and help establish an empirical foundation for political science researchers to explore with rigorous tools from causal inference frameworks.   In this paper, we propose a new scalable online Tensor Latent Dirichlet Allocation (TLDA) method with an end-to-end GPU implementation, ideally suited for the analysis of large text datasets. In summary, \textbf{we make the following contributions}:\\

\begin{itemize}
    \item \textbf{LDA with  Theoretical Foundations:} The method has \textit{identifiable and recoverable} parameter guarantees and \textit{sample complexity}  guarantees for large data. These theoretical properties provide assurance that under the assumed data generation process and mild regularity assumptions, the method returns accurate results in large data. 
    \item \textbf{Political Science Data Discovery:}
Our method improves understanding of large corpora at scale and answers questions in real-time about politically salient topics, such as the \#MeToo movement and social media activity around the Presidential election in 2020.  
\item \textbf{Fully Online, Incremental Tensor LDA:}
Our method can be estimated in real-time without relying on a precomputed dimensionality reduction of the $2^{\text{nd}}$-order moment. This results in a method that is computationally and memory efficient.
\item \textbf{Scaling to large corpora:} We demonstrate that the method scales linearly  by applying our approach to over 1 billion documents, scaling results which we show in the online appendix.
\item \textbf{Efficient implementation with end-to-end GPU acceleration}: In addition to our theoretical contributions, we release a new open-source library alongside this paper, which provides an efficient GPU-based implementation of all steps of topic modeling from pre-processing to tensor operations, without costly GPU-CPU exchanges.\footnote{The package is available at https://tensorly.org/tlda/dev/.} 
\end{itemize}

\begin{figure*}[th!]
    \centering
    \includegraphics[width=0.9\textwidth]{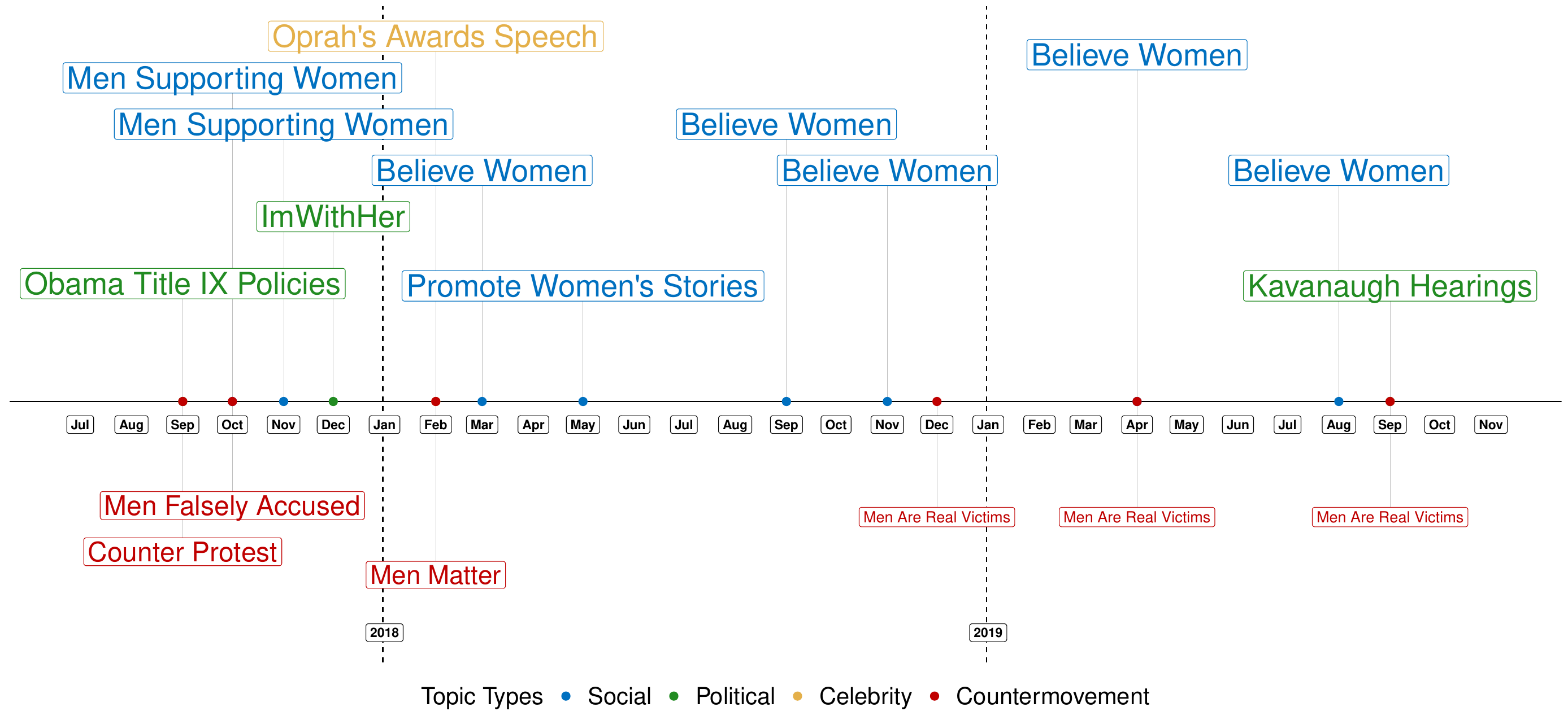}
     \caption{Evolution of the most prominent pro- and counter-movement topics in the \#MeToo discussion. In each iteration of the dynamic analysis described in Section \ref{ssec:qualitative}, we inspect the topics and manually label them, as well as classify them as pro- or counter- \#MeToo. We then display the topic in each category with the highest weight $\alpha_i$ below.}
    \label{fig:dyn_topic_evo}
\end{figure*}

To demonstrate the usefulness of our method, we utilize it to analyze the topical development of $8$ million tweets concerning \#MeToo from September 2017 through December 2019. Using our TLDA method, we can discern clear topical evolution over time through a qualitative study.
Notably, as we dynamically grow the corpus of tweets, we find that topics related to politically salient news events are generally ephemeral. In contrast, the topical prominence related to personal testimonies, coordinating protests, and supporting other participants in the \#MeToo movement stays persistently prominent as the topics evolve over time. In addition, discussion around counter-\#MeToo topics declined in prominence over time, and this discussion was subsumed into one topic by September 2019. (See Figures~\ref{fig:dyn_topic_evo} and \ref{fig:dyn_topic_pol_evo}). In addition to the applications mentioned here, our method can be used to study a wide array of political science questions where large, unstructured data could provide powerful evidence towards extant theories. Providing empirical baselines could also help inform the generation of new theories.

\section{Topic Models Continued Usefulness in Political Science Research}

Given the emergence of proprietary large language models (LLMs) and generative AI (ChatGPT, Claude), we demonstrate in this paper that online topic model methods provide several advantages to political science researcher, such as being theoretically founded, open-source, and scalable. We emphasize that our methods are designed for data discovery, establishing new empirical facts, and helping to clarify new theoretical frameworks, especially for data that are unstructured. That said, we of course caution applied researchers from blindly applying this (or any) method. We encourage readers to view this method as an important first step to lead to deeper analysis, connecting to additional datasets, and informing new theories, especially where existing data lacked either granularity or real-time dynamics.

\subsection{TLDA Provides Theoretical Foundations, Open Sourced Software, and Scalable Estimates}

Our method has three key advantages over existing methods.  First, our method has important theoretical properties. Neither the most popular LDA approaches based on \cite{blei2003} nor LLMs have yet the be shown to have these theoretical foundations \cite{Anandkumar2012,anandkumar2013spectral,JMLR:v15:anandkumar14b}. Our approach recovers -- in feasible computational time -- a provable identification guarantee for the topic-word probabilities and sample complexity bounds, as well as a form of statistical consistency in large samples \cite{Anandkumar2012,anandkumar2013spectral}.\footnote{Theorem \href{https://arxiv.org/pdf/1204.6703}{4.3} in \cite{anandkumar2013spectral} establishes the that spectral decomposition of the third order tensor - and thus our method -  accurately recovers topic-word probabilities, has no false positives, and and accurately recovers the underlying topics in the data. Theorem \href{https://arxiv.org/pdf/1204.6703}{5.1} in \cite{anandkumar2013spectral} establishes finite sample complexity guarantees, a form of statistical consistency. In the case of TLDA, we have $l_2$ norm guarantees for each column of the topic-word probability matrix.  Remark 11 in \cite{anandkumar2013spectral} notes that $l_1$ convergence, which is most familiar to political scientists, is not achievable in general. However, in the case that words are not uniformly distributed within topics (and the word-probabilities are concentrated in a small number of words with in a topic), $l_1$ convergence can be achieved.} These foundations suggest that parameter recovery  and large-sample accuracy are achievable when the text data follow a data generation process assumed by LDA, and that the topic-word probability matrix is full rank.  

Second, our method is open-source, intentionally designed to be estimated on a wide array of workstations. Although LLMs show tremendous promise as a research method and clearly model human speech patterns more realistically than bag-of-word approaches like we propose,  most of the popular LLMs are not open-source, cannot be trained locally without high-end computational resources \citep{Linegar2023}, and demonstrate various social biases with significant implications for research using their model outputs, particularly on prompts concerning gender and race \citep{bartl2020,kocielnik2023,nozza-etal-2021-honest,sheng-etal-2019-woman, zhao2018gender}.  By open-sourcing the training of the model, researchers can better test the sensitivity of their hyperparameter choices, run more robustness checks, and better assess the validity of their model outputs. While LLMs are extremely flexible and powerful and can be fine-tuned to a diverse array of tasks, this flexibility comes at the cost of unconstrained, high-dimensional parameter space: the prompts needed to define the task that the model will perform. Small changes in these prompts often result in large and unpredictable differences in the model outputs, which make prompt-based methods very challenging to tune and reduce their reliability. The limiting principles of prompt engineering are an area of active research. 

At the same time, in terms of interpretation and inference, proprietary LLM model weights, trained parameters, and specifics of training data are hidden from public view.  For commercial LLMs, the model underpinnings and mechanics are areas of active research by those developing the LLMs.  While it is possible for researchers to fix a particular version of an LLM to use in their research, it's not transparent to users how fixed any particular version of a commercial LLM might be. Unlike some proprietary models, our mathematical underpinnings are clearly and openly communicated and replicated, and the implementation itself is transparent and open source -- previous versions are archived and available for reproduction purposes. Relatedly, due to their generative objective, commercial LLMs have been shown to ``hallucinate'', to literally shift their responses to prompts to completely different topics and subjects; by construction our method will not hallucinate.  
Finally, by implementing a batched, streaming version of our method, researchers are freed from memory constraints that otherwise plague traditional unsupervised methods with large text corpora. 

Third, both LLMs and supervised methods impose substantive financial resource hurdles that serve as barriers to their use by many researchers -- hand labeling is expensive and proprietary LLM API calls can be extremely costly at scale. Furthermore, in applications where the entire population of documents contains valuable information, down-sampling may not be a viable solution to this cost. For example, ten thousand documents in a two-hundred million document sample (0.005\%) might comprise a cogent and important topic, say congressional speeches opposing war, but due to sampling, such a critical topic may not be identified at all if only a few thousand documents are sampled. For both replication purposes and ready-access to applied researchers, we hope our method allows political scientists to more readily answer questions of pressing concern, making more widespread use of large-scale text corpora numbering in the millions and billions of documents.

\section{Building on Methodological Innovations in Political Science and Computer Science}

In this section, we explain how we build from the foundation of two existing literatures. In the first case, we will leverage insights to build upon existing popular methods for topic models in political methodology. Second, we will contribute to a robust computer science literature on scalable LDA and Tensor LDA methods by proposing an GPU end-to-end pipeline for accurate and scalable open-source topic modeling. 

\subsection{Theoretical Guarantees Build on Existing Political Science Methods}

Our contribution to the political methodology literature is to introduce topic model techniques from computer science that have statistical theoretical foundations. To cluster and analyze large text datasets, political science researchers make widespread use of unsupervised topic models, which do not always have parameter recovery and accuracy guarantees \citep{anandkumar2013spectral,Anandkumar2012}. Among them, a popular model is Latent Dirichlet Allocation (LDA)~\citep{blei2003,hoffman2010}. This workhorse model can extract important information without requiring labeling or prior knowledge and has been used to analyze datasets across the social sciences, including studies on coordination among social movements, strategic communication of political elites, news dissemination, and the detection of toxic online behavior \citep{lauderdale_clark2014, HopkinsKing2010, grimmeretal2022}. Also popular is the Structural Topic Model (STM), which is closely related to LDA~\citep{roberts2014, roberts2016}. These methods are popular due to the feasibility and easy implementation, but they often do not scale well to large datasets, so we turn to the computer science literature where scale is key to unlocking new frontiers in research.

\subsection{Feasible Improvements for Existing Scalable LDA}

There have been numerous efforts to make LDA more scalable. Specifically, \cite{yu2015distributedLDA} develop a method for faster sampling and distributing computation across multiple CPU cores. More recently, efficient GPU LDA implementations have been proposed: some have developed improved GPU workload partitioning \citep{xie2019CuLDA, wang2020ezLDA}, while \cite{wang2020ezLDA}  developed a new branched sampling method. However, all of these implementations rely on traditional LDA methods based on Gibbs sampling, variational Bayes, or expectation maximization. The parallelization and scalability of these methods are inherently algorithmically challenging, as they are limited significantly by the sampling required to estimate the topics. Furthermore, unlike our method, these implementations are not available open-source, so they cannot be used easily by applied researchers.

Therefore, instead of relying on traditional LDA methods, we leverage tensor methods~\citep{kolda2009tensor}, which are embarrassingly parallel and have been proposed in order to scale to larger datasets~\citep{sidiropoulos2017tensor,papalexakis2016tensors}. 
Using tensor methods, it is possible to learn latent variables models with accuracy guarantees under mild regularity conditions~\citep{JMLR:v15:anandkumar14b,janzamin2019spectral}. In particular, Tensor LDA relies on the computation of third-order moments, i.e., the three-way co-occurrence of words, and decomposes them to recover the topics. This approach has been shown to have similar performance as traditional LDA~\citep{online}, which we also demonstrate empirically in the results of this paper.

However, previous implementations of these tensor methods are limited by (1) the explicit construction of the second- or third-order moments and (2) a lack of hardware acceleration due to being developed either entirely for CPU or with a costly exchange between CPU and GPU. In particular, researchers have developed tensor LDA methods that face memory constraints due to the computation of high-dimensional low-order tensors~\citep{anandkumar2013spectral, Anandkumar2012,JMLR:v15:anandkumar14b}. Further, these methods only run on CPU and do not benefit from hardware acceleration on GPU. But~\citet{JMLR:v16:huang15a} developed a stochastic tensor gradient descent (STGD) approach to estimate the third-order decomposition, allowing for further scaling of the method. However, this method has CPU-GPU exchange and relies on the explicit construction of the second-order moment, so it cannot be run fully online. Similarly, \cite{swierczewski2019large} proposed a method for learning the third-order decomposition using alternating least squares but is limited by a CPU-based implementation.

By contrast, we derive an efficient centered, online version of the TLDA that scales linearly to any dataset size (with constant memory).  We provide evidence supporting this claim in Table~\ref{tab:tlda_time_EF}. We provide an end-to-end GPU accelerated implementation that will be open-sourced along with this paper to enable its application to any dataset by other researchers. 

\begin{table}[t]
\caption{Runtime of our TLDA method on GPU for 260 million and 1.04 billion documents using the COVID dataset. None of the previous LDA methods scale to billions of documents.}
\label{tab:tlda_time_EF}
\centering
\begin{tabular}{ccc}
\toprule
\textbf{Model} & \multicolumn{2}{c}{\textbf{Time for Each Dataset (hours/minutes)}} \\
\cmidrule{2-3}
  & \textbf{260 Million docs} & \textbf{1.04 Billion docs}        \\
\midrule
Online TLDA & 3h28.2 & 13h09   \\
\bottomrule
\end{tabular}
\end{table}

\section{How to Achieve Scalable Tensor LDA}
\begin{figure*}[!ht]
    \centering
    \includegraphics[width=0.9\textwidth]{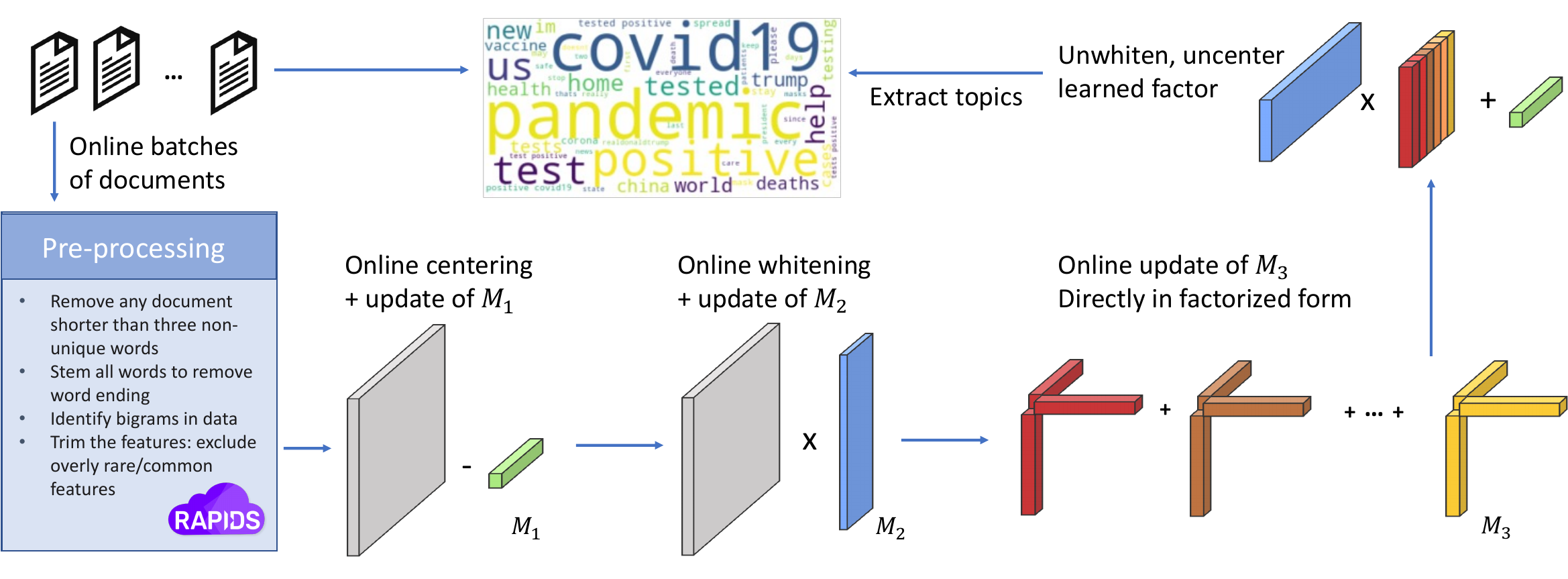}
        \caption{Overview of our approach. As batches of documents arrive, incrementally, they are first pre-processed (they are stemmed, tokenized, and the vocabulary is standardized). We then create a dataset of the counts for each word in each document. We then find the average number of times each word appears in each document (the average word occurrence, which is the first moment $M_1$) and subtract the value of $M_1$ from our existing word-frequency matrix. The resulting document term matrix is our centered dataset, $X$ ~(Section~\ref{ssec:m1}). We then perform a singular value decomposition on the centered data, $X$, to recover whitening weights without ever needing to calculate $M_2$, directly. This saves computationally overhead, while being mathematically equivalent. We then use these whitening weights to transform the centered data, $X$, which can be done incrementally~(Section~\ref{ssec:m2}). Finally, we construct the whitened equivalent of the third order moment, $M_3$, which is updated, directly in this factorized form~(Section~\ref{ssec:m3}). This learned factorization can be directly unwhitened and uncentered to recover the classic solution to TLDA~(Section~\ref{thm1}) and recover the topics and their associated word probabilities~(Section~\ref{ssec:postprocess}).}
    \label{fig:overview}
\end{figure*}

We provide a summary overview of our method in Figure~\ref{fig:overview}. As documents are provided as input to our TLDA pipeline, they are first pre-processed and converted into bag-of-words vectors. The key to TLDA methods is to perform no more dimension reduction than needed to ensure the method has theoretical accuracy guarantees, but to perform enough dimension reduction to achieve scale. \cite{anandkumar2013spectral} demonstrated two dimension reductions will produce accurate results. Similarly, our method takes the document-topic frequency matrix and takes the average word frequency across documents. We call this average $\matrix{M}_{1}$, the first moment. Then, we can demean the data and by doing so, we automatically update the model as we stream in new documents. Demeaning the data is a very powerful tool for reducing computational complexity because it cancels out very ugly off-diagonal terms for the higher-order moments that we calculate in our model. By canceling out these terms, we can now stream data into the method and automatically update the results.  Computer scientists call this an online update, terminology we adopt here and throughout the text (online centering and update of $\matrix{M_1}$ in Figure \ref{fig:overview}).  We then perform our first dimension reduction on the word co-occurrence matrix $\matrix{M_2}$ to find singular values, so we can use them to transform (``whiten") the data. Whitening is a linear transformation that reduces the dimensionality of the data from the number of words to the number of topics.\footnote{Whitening is a linear transformation that produces a new data matrix where each column is a de-correlated topic and the variance is standardized to $1$.} The word-occurrence matrix simply records how many times words occur together in the corpus.  Then, rather than directly compute the singular values for the word-occurrence matrix to get the whitening values, we can implicitly calculate them by performing PCA on the demeaned (centered) data, $\matrix{\tilde{x}}$ (online whitening and update of $\matrix{M_2}$ in Figure \ref{fig:overview}). Now, we take the whitening values and use them to transform the demeaned (centered) data. This ``whitening" step reduces the size of the data from $V\times V \times V$ (number of words by number of words) to $K \times K \times K$, the number of topics. This vastly reduces the size of the problem because every real-world application, the number of topics will be vastly smaller than the number of words. Having performed this transformation, our data have been reduced from size $V\times M$ to size $K \times M$.  We now compute the ``whitened" analogue of the word tri-occurrence matrix, $\matrix{M_3}$, which is $K \times K \times K$. We then find the eigenvalues of this $K \times K \times K$ object, which after some algebraic transformations, we show are equivalent to the topic model outputs from LDA (the unwhitened, uncentered learned factor in Figure \ref{fig:overview}).

Here, we propose several improvements over previous work to enable scaling the TLDA to billions of documents.  First, we reduce the computation complexity of the second and third-order cross moments; we derive them on centered data. Second, we incrementally estimate PCA on the centered documents. We take these principal components and use them to implicitly form the decomposition of the second order moment.\footnote{We leverage an algebraic relationship between principle components of the first moment and the SVD on the second moment to extract the whitening matrix without needing to fully calculate the SVD of the second order moment, which is computationally taxing. This implicit calculation allows us to parsimoniously decompose the data. This is helpful because the uncentered data are no longer sparse,  increasing the memory footprint of the dataset.} We derive a simplified, batched gradient update, leading to efficient recovery of the decomposition of the third-order moment. We jointly learn all moments online by updating the mean, PCA, and third-order decomposition in one pass, instead of relying on a precomputed dimensionality reduction of the second-order moment.

Crucially, we prove that the original topics computed in prior spectral LDA work can be recovered from the topics produced by our method with a simple post-processing step. As a result, our method enjoys the theoretical benefits of spectral LDA while significantly reducing its computational complexity. After finding the topic distribution over words, we employ standard variational inference to recover document-level parameters. We propose an efficient implementation on GPU within the TensorLy framework, which can easily scale to very large datasets.

\begin{figure*}[t]
    \centering
    \includegraphics[width=0.9\textwidth]{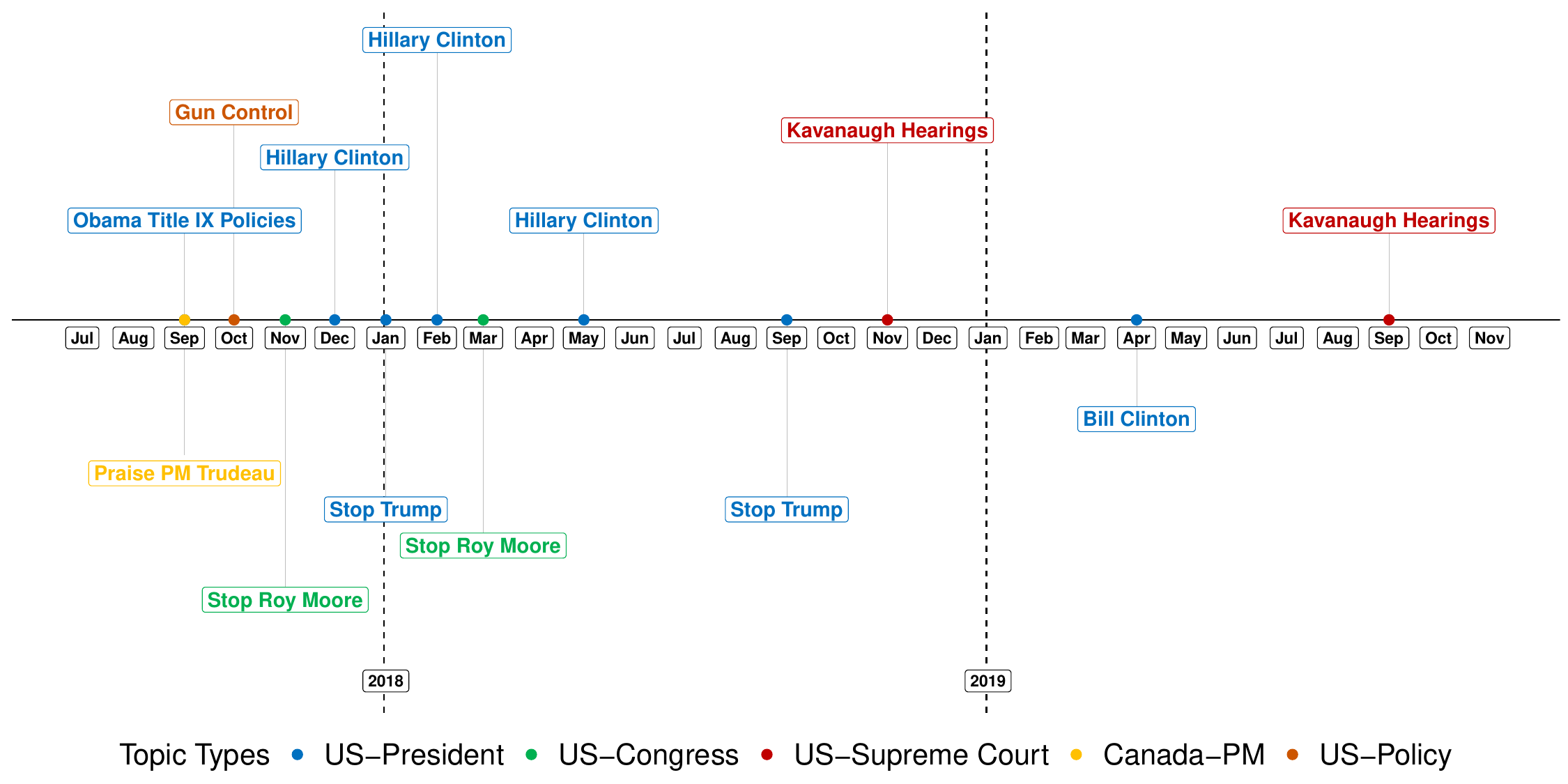}
        \caption{Evolution of most prominent political topics in the \#MeToo discussion. In each iteration of the dynamic analysis detailed in Section \ref{ssec:qualitative}, we inspect the topics, manually label them, and classify them as political or not political. We display the political topic with the highest weight $\alpha_i$ below.}
    \label{fig:dyn_topic_pol_evo}
\end{figure*}

\subsection{Computing the Centered Cumulants}\label{ssec:m1}

In this section, we provide a technical overview of the method. For a summary of our notation and the mathematical background for the method, please refer to the online appendix  (Table~\ref{tab:notation}). We begin by introducing the model, the data generation process, and the estimation routine.  

The data are a corpus of documents which we note in the following way. Let the data be a document term matrix $\matrix{F}$ with rows
$\vector{f}_t := (f_{t, 1}, f_{t, 2}, . . . , f_{t, V}) \in \R^{V}$ denoting the vector of word counts for the $t-$th document where $V$ is the number of words in the vocabulary, and let $N$ be the number of documents. Finally, we will let $K$ denote the total number of topics and $h$ be the topic labels. Then, the first order moment is:
\begin{equation}
    \vector{M}_1 := \frac{1}{N}\sum_{i=1}^N \vector{f}_i
\end{equation}

Our first innovation is to center the data by forming
\( %
    \tilde{\vector{x}}_i :=  \vector{f}_i - \vector{M}_1. 
\) %
Given this set-up, we can simplify the usual moments for spectral LDA (see \cite{Anandkumar2012,anandkumar2013spectral,JMLR:v15:anandkumar14b}, as well as Appendix \ref{ssec:uncentered-cumulants}). This is because the diagonalization terms in $\tensor{M}_2$ and $\tensor{M}_3$ become $\tensor{0}$ in expectation, as the first moment is now $\vector{0}$ for the centered data matrix $\tilde{\matrix{X}}$.   This vastly reduces the number of off-diagonal calculations required to estimate the higher-order moments. 

By removing the corresponding terms that are now $\tensor{0}$ in expectation from the expression of the moments in~\cite{anandkumar2013spectral}, we now have the following simplified empirical moments for the centered data $\tilde{\tensor{X}}$, where $\alpha_{0}$ is the topic mixing parameter.
\begin{eqnarray}
\tilde{\tensor{M}}_2 &:= &\frac{(\alpha_{0} +1)}{N}\sum_{t=i}^N \tilde{\vector{x}}_i \otimes \tilde{\vector{x}}_i \\
\tilde{\tensor{M}}_3 &:= &\frac{(\alpha_{0} + 1)(\alpha_{0} + 2)}{2N}\sum_{i=1}^N \tilde{\vector{x}}_i \otimes \tilde{\vector{x}}_i \otimes \tilde{\vector{x}}_i,
\end{eqnarray}
where $\otimes$ is the Kronecker product. 

Following~\cite{anandkumar2013spectral},
we know that the moments for the centered data can be factorized as:
\begin{eqnarray}
\expectation{}{\tilde{\vector{M}}_1} &:=& \vector{0} \\
\expectation{}{\tilde{\matrix{M}}_2} &:=& \sum_{i=1}^K \frac{\alpha_{i}}{\alpha_{0}} \vector{\vector{\nu}}_i \otimes \vector{\vector{\nu}}_i \\
\expectation{}{\tilde{\tensor{M}}_3} &:=&  \sum_{i=1}^K \frac{\alpha_{i}}{\alpha_{0}} \vector{\vector{\nu}}_i \otimes \vector{\vector{\nu}}_i \otimes \vector{\vector{\nu}}_i,
\end{eqnarray}

\noindent
where $\tilde{\vector{M}}_1$, $\matrix{\tilde{M}}_2$ and $\tilde{\tensor{M}_3}$ are the first, second, and third moments of the centered data, and $\matrix{\nu} = [\vector{\nu}_1, . . . , \vector{\nu}_K]$, the learned decomposition of $\tilde{\tensor{M}}_3$. Note that we use the centered analog from~\cite{anandkumar2013spectral}, which showed that the singular value decomposition of the third-order moment tensor yields estimates for the LDA model parameters. 

Finally, to recover the actual topic-word probability matrix as derived in ~\cite{anandkumar2013spectral} using the topics computed from the centered data, we prove Theorem~\ref{thm1}, demonstrating that the ground-truth factors can be recovered by de-centering the factors from the centered ones.

\begin{theo}
\label{thm1}
\vspace{-5pt}
Given the factors $\vector{\vector{\nu}}_i$ learned from the centered data, we  show that $$\mathbb{E}[\tensor{M}_3] = \sum_{i = 1}^K \frac{\alpha_i}{\alpha_0} (\vector{\nu}_i + \matrix{M}_1) \otimes (\vector{\nu}_i + \matrix{M}_1) \otimes (\vector{\vector{\nu}}_i + \matrix{M}_1).$$ That is, the true third-order cumulant can be recovered directly by re-centering the factors of the decomposition of the centered cumulant, indicating that the vectors $\vector{\nu}_i + \matrix{M}_1$ are equivalent to the ground-truth factors from ~\cite{anandkumar2013spectral}.
\end{theo}
\vspace{5pt}

For the proof, see Section~\ref{ssec:proof} in the online appendix.

Following from Theorem \ref{thm1}, we have
\begin{eqnarray*}
   \expectation{}{\tensor{M}_3} &=&  \sum_{i = 1}^K \frac{\alpha_i}{\alpha_0} (\vector{\vector{\nu}}_i + \vector{M}_1) \otimes (\vector{\vector{\nu}}_i + \vector{M}_1) \otimes (\vector{\vector{\nu}}_i + \vector{M}_1) \\
   &=& \sum_{i = 1}^K \frac{\alpha_i}{\alpha_0} \vector{\mu}_i \otimes \vector{\mu}_i \otimes \vector{\mu}_i,
\end{eqnarray*}
where $\vector{\mu} = [\mu_1, . . . , \mu_K]$ and $\vector{\mu}_i = Pr (f_j | h = i)$ where $\vector{h}$ are the set topic labels, $i$ is the $i$'th topic and $j$ is the $j$'th word in the vocabulary. In other words, $\vector{\mu}$ is the topic word matrix of the uncentered data $\matrix{F}$.

\subsection{Batched Implementation Enables Arbitrary Scaling}

In the subsections below, we present a batched implementation for TLDA. Each moment is individually calculated incrementally as we feed in data to the method, and then we estimate the topic-word probabilities using stochastic gradient descent.

In the section that follows the online decomposition for the second and third moments, we present a fully online implementation, which we recommend for very large datasets on the scale of billions of documents. For such datasets, the individual contribution of one data point is extremely small, so we average over many documents with minimal loss to accuracy given the enormous gains to scale.\footnote{For datasets on a smaller scale, individual documents can have much larger implications for topical inference, so the batched version is preferred, as empirical tests show that multiple loops through the data are necessary to meet the convergence criteria for the third order moment and for accurate inference}  In practice, this means we can iterate through the documents just once to still achieve our convergence criteria and to achieve accurate inference. As a byproduct of this implementation, we can then update the model by streaming new data points into the training API, giving a means to offer a fully online version of the model as part of this library.

\subsection{Online Decomposition of the Second Moment}\label{ssec:m2}
As a function of centering the data, our method streamlines the pipeline that~\cite{Anandkumar2012} proposes for calculating the second moment and the whitening matrix. Specifically, instead of constructing $\matrix{\tilde{M}}_2$, which is very memory-intensive for large data, we implicitly form $\matrix{\tilde{M}}_2$ by computing a singular value decomposition of the centered data matrix.

Using the singular values and singular vectors from the centered data, we construct a whitening matrix $\matrix{W}$ such that 
\begin{equation}
\matrix{W}\T \matrix{\tilde{M}}_2 \matrix{W}=I.
\end{equation}

We let $D$ be the whitening dimension size. We note that from~\cite{Anandkumar2012,anandkumar2013spectral,JMLR:v15:anandkumar14b}, letting $D = K$ is sufficient to compute the third-order decomposition, although a slightly larger $D$ can be chosen to improve the dimensionality reduction. Then, from the centered data, we have:
$$\matrix{W} = \frac{\sqrt{\alpha_0+1}}{N} \matrix{U}\, \matrix{\Sigma}^{-\frac{1}{2}},$$
where $\matrix{U}$ and $\matrix{\Sigma}$ (the variance matrix of the centered data) are the top $D$ singular vectors and singular values of the centered data, obtained through computing the PCA of $\matrix{\tilde{X}}$, which is equivalent to its SVD since the data is centered.

Then the $\matrix{\tilde{M}}_3$ tensor is implicitly formed using the whitened counts of the centered data. Whitening  renders the tensor symmetric and
orthogonal (in expectation). Most importantly, it reduces the dimensionality of the third moment from size $N^3$ to $D^3 \approx K^3$, where $K$ is the number of topics. Given the nature of speech in social environments, the number of topics will almost always be at least an order of magnitude smaller than the number of words.

To estimate the implicit third moment, the method calculates the whitened counts
$\matrix{x} = \matrix{W} \matrix{\tilde{X}}.$
We will use these whitened counts to construct the implicit third-order tensor. Using that implicit tensor, the method utilizes stochastic gradient descent to find the spectral decomposition of the third-order moments.

\subsection{Online Learning of the Third Moment}\label{ssec:m3}

Here, we formulate the TLDA framework in a vectorized form, solving for a \emph{batch} of data, resulting in a much more efficient implementation.  Let $\vector{\Phi} = [\Phi_1|\Phi_2| . . . |\Phi_K]$ be the eigenvectors of the third-order moment for the whitened, centered data. We note that each eigenvector $\Phi_i$ is of length $D$ and denote the full sample size as $N$. 

Now, note that the decomposition of the third-order moment for the whitened, centered data $\matrix{X}$ (of size $D \times D \times D \approx K \times K \times K$) is
$$\matrix{T} = \sum_{i \in K} \vector{\vector{\Phi}}_i  \otimes \vector{\vector{\Phi}}_i \otimes \vector{\vector{\Phi}}_i.$$

With the whitened tensor in hand, the method follows~\cite{JMLR:v16:huang15a} in implementing a batched Stochastic
Tensor Gradient Descent (STGD) algorithm for tensor CP decomposition.

Specifically, we consider a mini-batch of $n_B$ centered and whitened samples $\vector{x}_1, \cdots \vector{x}_{n_B}$, which we collect in a matrix $\matrix{X} \in \R^{n_B \times D}$. We want to learn a tensor factorization of the third-order whitened and centered cumulant:
\begin{align}
    \tilde{\tensor{M}}_3 = & \,\, \frac{(\alpha_0 + 1)(\alpha_0 + 2)}{2N} \sum_{n=1}^N \vector{x}_n \otimes \vector{x}_n \otimes \vector{x}_n\nonumber
    \\
    = & \,\, \frac{(\alpha_0 + 1)(\alpha_0 + 2)}{2N} \sum_{n=1}^N \vector{x}_n \otimes^3
\end{align}

We are trying to learn a rank-$K$ CP factorization with factors $\vector{\Phi}$ of $ \tilde{\tensor{M}}_3$ such that $\tilde{\tensor{M}}_3 = \tensor{T} = \sum_{i=1}^K \factor_i \otimes \factor_i \otimes \factor_i  $. In other words, we solve the following optimization problem:
\begin{equation}\label{eq:full_vec_pb}
    \argmin_{\factor;\, ||\factor_i||_F^2=1}
    \underbrace{|| \tilde{\tensor{M}}_3 - \tensor{T} ||_F^2}_{\text{reconstruction loss}} + \underbrace{ \frac{(1 + \theta)}{2} || \tensor{T} ||_F^2}_{\text{orthogonality loss}}
\end{equation}

In plain words, we minimize the reconstruction loss while inducing orthogonality on the decomposition factors. This can be seen by noting that the factors (and therefore the rank-1 components) are normalized, meaning the Frobenius norm of the second term simplifies to only the inner product between the components.

The problem in equation~\ref{eq:full_vec_pb} thus simplifies to:
\begin{align}\label{eq:vec_pb}
    \argmin_{\factor;\, ||\factor_i||_F^2=1} &
    \frac{(1 + \theta)}{2} || \sum_{i=1}^K \factor_i \kron^3 ||_F^2
    \\ \nonumber
    - & \inner{\sum_{i=1}^K \factor_i \kron^3}{ \frac{(\alpha_0 + 1)(\alpha_0 + 2)}{2N} \sum_{n=1}^N \vector{x}_n \otimes^3 }
\end{align}

This can be equivalently written in matrix form using the Khatri-Rao product:
\begin{align}\label{eq:vec_pb_equiv}
    \argmin_{\factor;\, ||\factor_i||_F^2=1} & \,\,
    \frac{(1 + \theta)}{2} || \factor \left(\factor \kr \factor\right)\T ||_F^2
    \\ \nonumber
    & -
   \frac{(\alpha_0 + 1)(\alpha_0 + 2)}{2n} \inner{\factor \left(\factor \kr \factor\right)\T}{ \matrix{X}\T \left( \matrix{X} \kr \matrix{X} \right) }
\end{align}

By taking the derivative with respect to the decomposition factor $\factor$, we get:
\begin{equation}\label{eq:vec_deriv}
\myd{\mathcal{L}}{\factor} =
3(1 + \theta) \factor (\factor\T\factor * \factor\T\factor) - \frac{3(\alpha_0 + 1)(\alpha_0 + 2)}{2n} \matrix{X}^T (\matrix{X}\factor*\matrix{X}\factor)
\end{equation}

We then update the factor via batched stochastic gradient update:
\begin{equation}
    \vector{\Phi}_{t+1} = \vector{\Phi}_{t} - \beta \myd{\mathcal{L}}{\factor},
\end{equation}
with $\beta$ the learning rate.

\subsection{Fully Online Implementation}

In a fully batched implementation above, the method relies on computing each higher-order moment sequentially, even though each moment is individually learned online. By contrast, in the fully online version of our TLDA method presented here, we learn both the moments by jointly learning both the second and third-order moments online.\footnote{The first moment is just the average word frequency and trivial to compute at scale.} We first find initial values for the factors by running the batched TLDA to convergence on a small portion of the data. Then, when given a new batch of data, we first update $\tensor{M}_1$, then update the incremental PCA (and use the new version of the PCA to whiten the data), and finally perform a gradient update of the third-order moment using the new batch of whitened data. As a result, we can update the third-order moment decomposition only using the new batch without looping through any prior data. This is in contrast to the batched version of our method, where we loop through the entire dataset three times to compute the first-order moment, second-order moment decomposition, and third-order moment decomposition, respectively.

Although we only perform one gradient descent step per batch of data in this version of the method, we expect that for large datasets, there are sufficiently many documents so that this does not significantly impact the quality of the factors produced.\footnote{We confirm this claim empirically in Section 5.} We use online LDA to obtain topic coherence values that are similar to or better than existing methods.

\subsection{Recovering the Topic Model Parameters}\label{ssec:postprocess}

Once we have learned the factorized form $\Phi$ of the third-order moment, we describe how we recover the uncentered, unwhitened moment to recover the topics.
First, we obtain $\matrix{\nu}$, the estimate of the decomposition of $\tilde{\matrix{M}}_3$, by unwhitening the components $\Phi$ of the decomposition:
$$\matrix{\nu} = \matrix{W}^{T^{\dagger}} \matrix{\Phi} $$
where $\dagger$ denotes the pseudo-inverse. Using this, we can find 
$$\alpha_i = \gamma^2\vector{\nu}_i^{-2}$$
Here, $\gamma$ is a scaling factor such that $\sum_{i=1}^k \alpha_i = 1$.

As derived in Theorem \ref{thm1}, we can then re-center $\matrix{\nu}$ to compute the estimate for the uncentered topic-word probabilities
$$\vector{\mu}_i = \vector{\nu}_i + \matrix{M}_1$$

\section{The Tensorly-LDA Package}
Along with this paper, we release a new Python package that provides an efficient, end-to-end GPU-accelerated implementation of our proposed online Tensor LDA.\footnote{The package installation instructions can be found on its website, here: \url{https://tensorly.org/tlda/dev/install.html}.} We used that implementation for all experiments in this paper. It consists of two main steps: an efficient pre-processing module that uses RAPIDS and a module that builds on top of TensorLy to learn the higher-order cumulants. 

Our entire Tensor LDA method is end-to-end GPU accelerated and implemented on the Nvidia Rapids Data Science Framework, a GPU-based architecture for data analysis in python~\citep{rapids}, as well as the TensorLy library, a high-level API for tensor methods in Python~\citep{tensorly}. First, the data is pre-processed, on GPU, using RAPIDS. After the data has been pre-preprocessed, all tensor operations are performed using the TensorLy library, which is used to learn the third-order cumulant in factorized form directly. RAPIDS is used for learning the second-order cumulant through incremental PCA. The result is an end-to-end GPU implementation of a large-scale topic model with no CPU-GPU exchange. We empirically establish our implementation in the next section through thorough experiments and demonstrate large speedups over previous works.

The library provides all the tools to run our method on actual dataset. To facilitate adoption by practitioners, it comes with a thorough online documentation and interactive examples. Both the examples and an extensive suite of unit-tests are run dynamically after any change in the codebase through a continuous integration suite to ensure correctness.

\subsection{Input and Output}

The method takes a data matrix where a  each entry is the centered word frequencies for each of $V$ words for $N$ document. Each column represents a word in the dataset and each row is a document, for a matrix of size $N\times V$. The method produces two key outputs: first, the topic-word matrix and second, the learned weights, $\alpha_i$. These outputs can be fed as inputs into a standard variational inference (VI) method that calculates the topic-document matrix, as well. We have included a standard VI method in our API so users can calculate the document-topic matrix for their applications. 

\paragraph{Hyperparameter Tuning}

There are several hyperparameters standard to LDA and STGD methods that under researcher discretion. Users are encouraged to check that the default parameters are appropriate for their application.

\begin{enumerate}
    \item Number of topics $k$: number of learned clusters. Should be optimized by researcher.        
    \item Topical mixing, $\alpha_0$: The level of mixing believed to be in the documents. Closer to $0$ is no mixing and closer to $\infty$ means fully mixed documents.  
    \item Learning rate $\beta$. How much to allow new batches of data to contribute to the factor update. Needs to be tuned for stable convergence (if convergence is too slow, increase it. If topics appear noisy or nonsensical, decrease it).
    \item Orthogonality penalty $\theta$: How much separation you expect between topics. If topical mixtures appear too similar, increase this parameter. If topics are incoherent or convergence is unstable, decrease it.  

\end{enumerate}

\paragraph{Recommendations for Data Pre-processing}
By pre-processing the data on the Rapids GPU framework, we alleviate a crucial bottleneck in the practicability of LDA on large datasets. Although pre-processing has been shown to be critical to producing valid results, especially in social science contexts~\citep{grimmer2013}, existing frameworks for topic models often entail expensive CPU-GPU exchange. Having overcome this bottleneck, we follow best practices suggested by \cite{grimmer2013} and summarized in \cite{HopkinsKing2010}; we optimize feature selection by stemming and tokenizing the data. The political science literature has found that for non-noisy inference on text data, we want neither too few common features such that there is no variation amongst documents nor too many uncommon features such that there are no distinguishable clusters.  

We follow this process to arrive at our final set of features:
\begin{itemize}
    \item Remove any document shorter than three non-unique words
    \item Stem all words to remove word endings using a Porter Stemmer 
    \item Identify bigrams in the data
    \item Trim the features: exclude any features appearing in fewer than the lower bound that scales with the number of documents of the document and more than the upper bound that scales with the number of documents.  
\end{itemize}

The political science literature has intensively explored the sensitivity of critical substantive findings to pre-processing. \cite{HopkinsKing2010} find that the consensus in the social science literature is that brute force unigram-based methods, with rigorous
empirical validation, will typically account for the majority of the available explanatory power in the data. So long as pre-processing captures all relevant features, our inferences derived from NLP can be used to analyze social phenomena.   However, \cite{HopkinsKing2010} note that the tuning of pre-processing choices generally depends on the nature of the application. In all of the applications presented in this paper, our unit of observation is a tweet, an inherently short document limited to 270 characters. Finally, following standard practice for topic models, we stem words to their base root so that the core meaning of these words is captured by only one token. In previous applications, this has been shown to both improve computational tractability and clarify the substantive analysis of text data.

\subsection{Package Availability}
The package is fully open-source as part of the TensorLy~\citep{tensorly} project, under BSD-3 license, which makes it suitable for any use, academic or industrial. It is well tested, has extensive documentation.

\section{Simulations}

\begin{table*}[htbp]
\centering
\caption{Comparison of Topic Recovery on Synthetic Data for Various TLDA Methods}%
\label{tab:tlda_cor_20k}

\begin{tabular}{c ccc ccc}
\toprule
\textbf{\begin{tabular}[c]{@{}c@{}}Vocabulary
\\ Size\end{tabular}} & \multicolumn{3}{c}{\textbf{Average Correlation}} 
& \multicolumn{3}{c}{\textbf{\begin{tabular}[c]{@{}c@{}}Standard
\\ Deviation of Correlation\end{tabular}}}                          \\ \cmidrule{2-7} 
                                                                        & \multicolumn{1}{c}{\textbf{\begin{tabular}[c]{@{}c@{}}PARAFAC\\ TLDA$^{1}$ \end{tabular}}} & \multicolumn{1}{c}{\textbf{\begin{tabular}[c]{@{}c@{}}SGD\\ TLDA$^{2}$ \end{tabular}}} & \textbf{\begin{tabular}[c]{@{}c@{}}Batched\\ TLDA$^{3}$ \end{tabular}} & \multicolumn{1}{c}{\textbf{\begin{tabular}[c]{@{}c@{}}PARAFAC\\ TLDA$^{1}$ \end{tabular}}} & \multicolumn{1}{c}{\textbf{\begin{tabular}[c]{@{}c@{}}SGD\\ TLDA$^{2}$   \end{tabular}}} & \textbf{\begin{tabular}[c]{@{}c@{}}Batched\\ TLDA$^{3}$\end{tabular}} \\
\midrule
500                                                                                 & \multicolumn{1}{c}{0.886}                                                           & \multicolumn{1}{c}{0.893}                                                              & \textbf{0.943}                                                   & \multicolumn{1}{c}{0.304}                                                           & \multicolumn{1}{c}{0.306}                                                              & \textbf{0.115}                                                   \\ 
1000                                                                                & \multicolumn{1}{c}{0.906}                                                           & \multicolumn{1}{c}{0.787}                                                              & \textbf{0.930}                                                   & \multicolumn{1}{c}{0.183}                                                           & \multicolumn{1}{c}{0.404}                                                              & \textbf{0.090}                                                   \\ 
1500                                                                                & \multicolumn{1}{c}{0.872}                                                           & \multicolumn{1}{c}{0.840}                                                              & \textbf{0.873}                                                   & \multicolumn{1}{c}{0.265}                                                           & \multicolumn{1}{c}{0.363}                                                              & \textbf{0.101}                                                   \\ 
\bottomrule
\end{tabular}
\parbox{6in}{Note: (1) \cite{JMLR:v15:anandkumar14b}; (2) \cite{JMLR:v16:huang15a}; (3) Our method.  }
\end{table*}

\begin{table}[htbp]
\centering
\caption{Comparison of CPU Runtime on Synthetic Data for Various TLDA Methods}%
\label{tab:tlda_time_20k}

\begin{tabular}{c ccc}
\toprule
\multicolumn{1}{c}{\textbf{\begin{tabular}[c]{@{}c@{}}Vocabulary\\ Size\end{tabular}}} & \multicolumn{3}{c}{\textbf{Average   Time (s)}}                                                                                                                                                                                                   \\ \cmidrule{2-4} 

                                                                                    & \multicolumn{1}{c}{\textbf{\begin{tabular}[c]{@{}c@{}}PARAFAC \\ TLDA$^{1}$\end{tabular}}} & \multicolumn{1}{c}{\textbf{\begin{tabular}[c]{@{}c@{}} SGD\\                     TLDA$^{2}$\end{tabular}}} & \textbf{\begin{tabular}[c]{@{}c@{}}Batched\\ TLDA$^{3}$ \end{tabular}} \\ \midrule
500                                                                                 & \multicolumn{1}{c}{20.00}                                                            & \multicolumn{1}{c}{53.87}                                                              & 3.00                                                             \\ 
1000                                                                                & \multicolumn{1}{c}{33.40}                                                            & \multicolumn{1}{c}{71.75}                                                              & 2.98                                                             \\ 
1500                                                                                & \multicolumn{1}{c}{62.18}                                                            & \multicolumn{1}{c}{103.39}                                                              & 3.55                                                             \\ \bottomrule
\end{tabular}
\parbox{6in}{Note: (1) \cite{JMLR:v15:anandkumar14b}; (2) \cite{JMLR:v16:huang15a}; (3) Our method.  }
\end{table}

\subsection{Parameter Recovery and Comparison to Previous TLDA Methods}
In this section, we demonstrate that our method results in gains in accuracy, topic correlation, and speed in comparison to existing TLDA methods in a simulated setting. We use the traditional LDA Data Generation Process for generating the simulated data. (See Appendix \ref{sec:simDGP} and \cite{blei2003}). We present a comparison to two key existing versions of the TLDA method: (1) The spectral decomposition algorithm in~\citep{anandkumar2013spectral,Anandkumar2012,JMLR:v15:anandkumar14b} and (2) the SGD-based TLDA derived in~\citep{JMLR:v16:huang15a}, in which the third-order moment is computed online. To do so, we show comparisons of all three TLDA methods on synthetic data. Due to the small scale of the synthetic experiment (20,000 documents), we run the batched version of our method. 

 As a result, through this process, we obtain ground-truth factors that adhere to the assumptions of LDA. By running TLDA on the synthetic document vectors, we can then compare each factor in the learned topic-word matrix to the corresponding factor in the generated ground truth topic-word matrix by computing their correlation.

However, the topics in the learned topic-word matrix can be in any order, so we limit the number of topics to $K = 2$ and use the permutation that maximizes the average correlation to the ground truth. We use parameters $\alpha_0 = 0.01$ and whitening size $D = 2$ for all TLDA methods, as well as learning rate $1 \times 10^{-4}$ for the SGD-based TLDA and our method. Table \ref{tab:tlda_cor_20k} shows the correlation between the learned and ground-truth factors in corpora with $20,000$ documents. The results are averaged over $10$ random seeds for each combination of parameters. This table illustrates that under a variety of vocabulary sizes, our method is more accurate than existing tensor methods, as evidenced by the higher mean and lower standard deviation of correlations among all runs. 

In Table \ref{tab:tlda_time_20k} we compare the average runtime of the three TLDA methods for the synthetic data experiments in Table \ref{tab:tlda_cor_20k}. In appendix \ref{tab:new_comp}, we also compare against other scalable LDA methods -- we note our TLDA method compares favorably and that none of the methods against which we compare have theoretical accuracy guarantees. We run this analysis on CPU due to the relatively small scale of the experiment. Our version of the TLDA method results in a runtime that is between 6 and 20 times faster than the existing TLDA methods. Furthermore, as shown in Table \ref{tab:tlda_time_20k}, as the vocabulary size increases, the runtime of our method increases its relative speed advantage over the others. This demonstrates the value of the simplifications we introduce in the method section; we significantly outperform the existing TLDA methods in terms of runtime while also making non-trivial gains in accuracy.

\section{Applications}
In this section, we demonstrate the scalability of our method by applying it to two large-scale Twitter datasets -- concerning the \#MeToo movement and the 2020 U.S. Presidential elections. These applications present analyses of  important datasets that political scientists might wish to use to study collective action, political behavior, gender politics, election misinformation, and many other theoretically- and substantively-important topics.  But analyses like these would have been infeasible or perhaps impossible due to the large size of the data (as we demonstrate below) without methods like TLDA. 

The \#MeToo dataset comprises 7.97 million tweets related to the political and social discussion surrounding \#MeToo. We conduct thorough ablation studies using the \#MeToo dataset and empirically demonstrate that the runtime of our method scales linearly as the number of documents increases and is near constant as the number of topics increases. We compare the runtime and topic coherence to a popular off-the-shelf model to show that our online method is 15-140x faster than previous methods while achieving similar or better coherence.\footnote{The machines used for all timing experiments are reported in Table \ref{tab:machines} in the appendix.}  Additionally, we show the practical utility of our method for applied researchers by using it to dynamically analyze the evolution of the \#MeToo dataset over time. We show qualitative evidence of topical evolution in the discussion around the social movement and political coordination on social media.

The second application uses a dataset of approximately 29 million tweets that were collected during the 2020 presidential election, regarding the conduct of the election. We show below that our method can process and analyze these data quickly and efficiently, generating interesting topic modeling results that could shed light on important political science questions. 

Finally, in the online appendix we analyze a third dataset that contains over 260 million tweets related to COVID-19, collected in real-time using keywords from the Twitter streaming API. We show that our method produces coherent estimates in under 3.5 hours on this dataset, and in under 13.5 hours on a simulated 1.04 billion document dataset created using these COVID-19 data. Thus, we demonstrate that our method is effective for unsupervised analysis of large-scale data on the order of billions of documents.

\subsection{The \#MeToo movement:  Scaling, Ablation and Substantive Studies}

\begin{table*}[!t] %
\caption{TLDA Convergence timing comparison on full \#MeToo dataset}
\vspace{-2pt}
\label{tab:gensim_time}
\centering
\begin{tabular}{cccccccc}
\toprule
\multicolumn{1}{c}{\textbf{\begin{tabular}[c]{@{}c@{}}Number \\ of Topics\end{tabular}}}
 & \textbf{Model} & \textbf{Time (s)} & \textbf{Speedup} & \multicolumn{4}{c}{\textbf{Coherence}}                                 \\
\cmidrule{5-8} 
                                                                                      &                                 &                                                                                           &                                                                                         & \textbf{$\text{C}_\text{v}$}  & \textbf{$\text{C}_{\text{UCI}}$} & \textbf{$\text{C}_{\text{NPMI}}$} & \textbf{$\text{U}_{\text{Mass}}$} \\
\midrule
10                                                                 & Gensim LDAMulticore                          & 1581.80                                                                                   & -                                                                                       & \textbf{0.421} & 0.481           & 0.061            & -4.172           \\
                                                                                      & Batched TLDA                    & 121.81                                                                                    & 12.99                                                                                   & 0.389          & \textbf{0.489}  & \textbf{0.063}   & \textbf{-3.514}  \\
                                                                                      & Online TLDA                     & \textbf{103.45}                                                                           & \textbf{15.29}                                                                          & 0.407          & 0.473           & 0.061            & -3.528           \\
                                                                                      \midrule
20                                                                  & Gensim LDAMulticore                         & 1528.38                                                                                   & -                                                                                       & \textbf{0.420} & \textbf{0.493}  & 0.058            & -4.514           \\
                                                                                      & Batched TLDA                    & 145.33                                                                                    & 10.52                                                                                   & 0.393          & 0.473           & \textbf{0.062}   & \textbf{-3.469}  \\
                                                                                      & Online TLDA                     & \textbf{96.10}                                                                            & \textbf{15.90}                                                                          & 0.370          & 0.440           & 0.056            & -3.561           \\
                                                                                      \midrule
40                                                                  & Gensim LDAMulticore                          & 8801.33                                                                                   & -                                                                                       & 0.404          & 0.479           & 0.053            & -5.075           \\
                                                                                      & Batched TLDA                    & 149.54                                                                                    & 58.86                                                                                   & \textbf{0.416} & \textbf{0.499}  & \textbf{0.066}   & \textbf{-3.474}  \\
                                                                                      & Online TLDA                     & \textbf{98.27}                                                                            & \textbf{89.56}                                                                          & 0.384          & 0.451           & 0.057            & -3.609           \\
                                                                                      \midrule
60                                                                   & Gensim LDAMulticore                           & 14448.87                                                                                  & -                                                                                       & 0.383          & 0.377           & 0.045            & -5.446           \\
                                                                                      & Batched TLDA                    & 152.67                                                                                    & 94.64                                                                                   & \textbf{0.402} & \textbf{0.459}  & \textbf{0.060}   & \textbf{-3.466}  \\
                                                                                      & Online TLDA                     & \textbf{126.19}                                                                           & \textbf{114.50}                                                                         & 0.380          & 0.433           & 0.055            & -3.622           \\
                                                                                      \midrule
80                                                                   & Gensim LDAMulticore                          & 14700.59                                                                                  & -                                                                                       & 0.382          & 0.324           & 0.042            & -5.434           \\
                                                                                      & Batched TLDA                    & 158.88                                                                                    & 92.53                                                                                   & \textbf{0.389} & \textbf{0.465}  & \textbf{0.061}   & \textbf{-3.488}  \\
                                                                                      & Online TLDA                     & \textbf{120.13}                                                                           & \textbf{122.37}                                                                         & 0.355          & 0.414           & 0.052            & -3.624           \\
                                                                                      \midrule
100                                                                  & Gensim LDAMulticore                          & 14830.54                                                                                  & -                                                                                       & \textbf{0.381}          & 0.352           & 0.043            & -5.176           \\
                                                                                      & Batched TLDA                    & 171.82                                                                                    & 86.31                                                                                   &       0.375         &    \textbf{0.439}             &        \textbf{0.058}          &         \textbf{-3.524}         \\
                                                                                      & Online TLDA                     & \textbf{103.45}                                                                           & \textbf{143.36}                                                                         & 0.355          & 0.417           & 0.052            & -3.647    \\
                                                                                      \bottomrule
\end{tabular}
\parbox{5in}{Note:  See \cite{rehurek2011gensim} for details about Gensim LDA Multicore and \cite{rolder_2015} for the coherence metrics.}
\end{table*}

\subsubsection{Studying mass movements and collective action with large-scale datasets} 

The \#MeToo movement is a prolific women's rights movement that gained traction extremely quickly on Twitter in October 2017, with over 7.9 million tweets containing the \#MeToo hashtag from October 2017 to October 2018 alone \citep{metoo_news}.  This movement is an important example of what Clark-Parsons has termed ``Network Feminism'', where social media platforms have become a crucial organizational tool for mobilization of social and political movements \citep{clarknetworked}. 

Going back to the early theoretical work of Mancur Olson, studying social movements and protests politics as a lens for collective action has been an important literature in political science \citep{olson1971logic}.  In particular, researchers have long tried to understand the political origins of protest politics and mass movements, because as Olson noted participation can be costly and the results of participation can be difficult for individuals to assess.  

Studying how mass movements and protest politics arise, how they are organized, and how they are sustained in the long run, is also complicated by a lack of available data.  Movements and protests arise quickly, authorities often act to stop and prevent protesting and organizing, which means in many cases that political scientists cannot often collect data about protests and movements. Surveys of protest participants can be done after the fact, but they can be difficult to find, difficult to persuade to participate in a survey, and their survey response may be inaccurate with the passage of time.  Thus, much of the literature has resorted to case studies of historical examples \citep{chong2014collective}.

In recent years, as the use of social media by protest and movement participants has sparked new research in about collective action, specifically regarding protests and mass movements.  By collecting data from participants in the protests, while they are protesting or acting collectively, has proven to be an important way to generate datasets to test existing theories, for example about the Arab Spring or Black Lives Matter movements \citep{steinert-threlkeld2017, kannetal2023}.  

In a similar way, the tweets related to \#MeToo thus provide rich data for investigating the evolution of a modern social movement initiated by online discussions on social media.   What topics engaged participants in the \#MeToo movement?  Did the topics of conversation change over the course of the movement?  Can the language of social media conversations help us determine the motivations of participants in the movement, where they motivated by self-interest or collective concerns?  These data can be crucial for understanding this important social and political movement, and for testing theories about how movements like these arise and are sustained.  

\begin{figure}[!ht]
    \centering
    \includegraphics[width=0.4\textwidth]{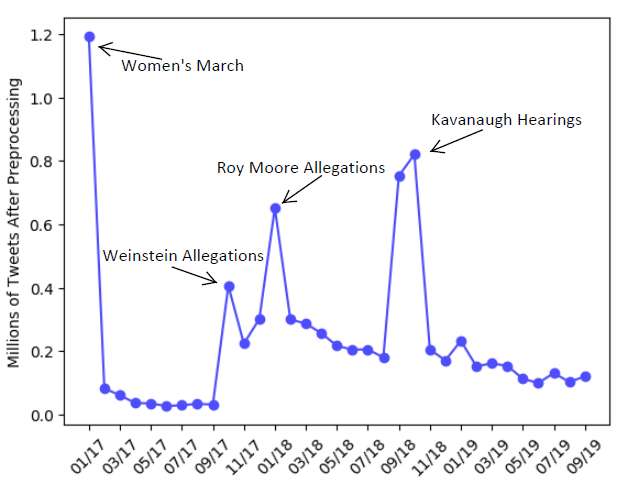}\vspace{-10pt}
    \caption{\textbf{Tweets per month in the \#MeToo data, in millions.}}
    \label{fig:n_tweets}
\end{figure}

We analyze the topics present in a corpus of \#MeToo tweets collected from January 2017 to September 2019, which contains 7.97 million tweets after pre-processing. Figure \ref{fig:n_tweets} shows the initial proliferation of tweets in January 2017 related to the Women's March and Movement, as well as the viral growth of the \#MeToo movement in September and October 2017.  

\vspace{-5pt}

\subsubsection{Pre-Processing}  We follow the standard pre-processing framework outlined earlier in the paper, which we note takes only 178 seconds on GPU using RAPIDS. The \#MeToo Twitter data are high-frequency, and there are orders of magnitude many more observations than in existing datasets analyzed in applications in the social science literature. Thus, we set a very low lower-bound for pruning words -- words need only appear in 0.002 percent of documents to be included in the \#MeToo data. Higher thresholds would dramatically reduce the number of features such that there was no variation in the document structure. Any lower and many features which only appear in a handful of documents would be essentially noise. These are proper nouns such as usernames, typos, nonsensical words, or words that are not common enough to help define meaningful clusters.  At the same time, we exclude words that appear in more than 50 percent of the documents. This only amounts to removing 50 words, many of which are so common as not to be useful in delineating a topic due to lack of variation (as they appear in every document).  We arrive at 1837 words, more than enough to pin down meaningful topics.
Changing this cutoff to 70, 80, or 90 percent does not significantly change the number of words in the vocabulary. Still, we stick to the more restrictive cutoff because many common words would otherwise dominate every topic, hindering the interpretability of the model.  We also ensure that words occur in at least 0.002 percent of documents, so idiosyncratic words that explain slight variation in the data are excluded. These words might appear in only one or two documents, which is far too infrequently to pin down substantive topics. Finally, following standard practice, we stem words by cutting off verb and noun endings so that base words will carry the same semantic meaning.

\subsubsection{\#MeToo Scaling and Convergence Speed Comparison}In this set of results, we first run the batched and online versions of our TLDA method on the entire \#MeToo Twitter dataset on one GPU core to analyze how quickly both versions of our approach converge with varying numbers of topics. We then compare the scaling of our TLDA method to that of Gensim by computing convergence time on subsets of the \#MeToo dataset containing 1M, 2M, 5M, and 7.97M documents.

\paragraph{Full \#MeToo timing comparison} To choose the optimal parameters for our method, we run an extensive grid search over the number of topics $K$, whitening sizes $D$, topic mixing parameters $\alpha_0$, and learning rates $\beta$. For each number of topics, we determine the optimal parameters by finding the parameter combination with the highest mean coherence score across all metrics. For $K = 10$, we report the 20 top words for the batched model with the best parameters in Table \ref{tab:metootopics} in the appendix. We include the parameters used for each number of topics in Table \ref{tab:tlda_params} in the appendix.

Using these optimal parameters, we run the batched and online versions of our model to convergence and provide a benchmark comparison of speed and coherence on GPU to the most popular off-the-shelf CPU-based LDA method, Gensim LDAMulticore, which is fully parallelized~\citep{rehurek2011gensim}. For the Gensim LDA model, we keep all default parameters, except we increase the number of passes and iterations until the coherence of the model converges for each specified number of topics. We include the final parameters used in Table \ref{tab:gensim_params} in the appendix. We compute the coherence measures using the Gensim CoherenceModel library by providing the $20$ top words in each topic and using the default parameters for each coherence measure.

In the appendix, we also compare how our methods scale against other state-of-the-art LDA methods (See Table \ref{tab:new_comp}). In particular, we analyze LightLDA \cite{Yuan2015} and WarpLDA \cite{chen2016}, both of which purport tremendous increases in scale. We note neither method offers the accuracy guarantees of TLDA, nor are they actively maintained as part of a larger suite of packages (such as Gensim's LDAMulticore or Tensorly's TLDA, presented here). We note in any case that our method outperforms both in terms of time, even without the theoretical guarantees of our method  \cite{anandkumar2013spectral}\footnote{We not that not only is our method faster than prior scalable LDA work, but it also allows for the benefits of TLDA, such as these convergence guarantees and exploiting third-order word co-occurrences, to be used for modeling large-scale data for the first time.}. We also note that there are many other methods purporting to perform topic modeling at scale. As far as our review of them can determine, none purport to have the accuracy guarantees of our method, nor have they been widely adopted. For this reason, we focus our analysis on the Gensim's implementation, which is, by far, the most popular and widely adopted.

\begin{figure}[!t]
    \centering
    \includegraphics[width=0.45\textwidth]{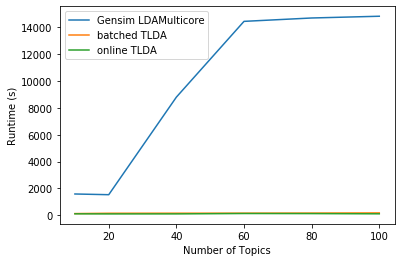}\vspace{-10pt}
    \caption{\textbf{Runtime comparison for TLDA on GPU vs Gensim for the full \#MeToo corpus and varying numbers of topics}. This shows that the runtime of our method scales near-constantly with respect to the number of topics, while Gensim scales more than linearly.}
    \label{fig:GPUabstudyfit-topics}
\end{figure}

As shown in Table \ref{tab:gensim_time}, we find that the batched version of our method, running on a single GPU, achieves 10 to 95x speedup over Gensim running in parallel on 79 CPU cores. The online version running on a single GPU achieves a 15-143x speedup over Gensim run in parallel on 79 CPU cores. Notably, our method has relatively constant convergence speed and consistent coherence for all numbers of topics. On the other hand, as the number of topics increases, the convergence time for the Gensim model increases nonlinearly and the final coherence score to which it converges decreases significantly. The runtime comparison as the number of topics increases is visualized in Figure~\ref{fig:GPUabstudyfit-topics}. As these results indicate, our method is especially useful for analyzing larger numbers of topics, thanks to the near-constant scaling of convergence time as the number of topics increases: for more than 60 topics our method is over 100x faster on GPU than the fully parallelized Gensim CPU LDA method.

\begin{figure}[!t]
    \centering
    \includegraphics[width=0.45\textwidth]{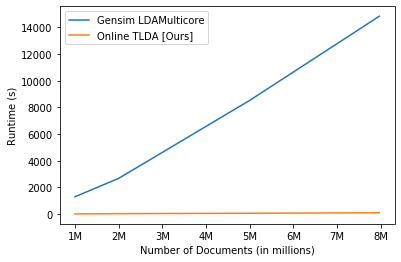}\vspace{-10pt}
    \caption{TLDA vs Gensim fitting time. We compare the time to fit Gensim's LDAMulticore and our online TLDA, not including pre-processing, for 100 topics. We plot the runtime in seconds as a function of the size of the subset from the \#MeToo dataset, from 1 million to 7.97 million tweets.}
    \label{fig:speedup}
\end{figure}

\paragraph{Scaling study} We fit our TLDA method for 10, 20, 40, 60, 80 and 100 topics on four subsets of the \#MeToo movement containing 1, 2, 5, and 7.97 million tweets. We keep the same vocabulary and optimal parameters from the full \#MeToo timing comparison while running the model on each of these subsets. As seen in Figure~\ref{fig:speedup}, which displays the TLDA fit time (excluding preprocessing) for 100 topics, convergence time for our method scales linearly with the number of documents on GPU, while convergence time for the Gensim LDAMulticore method increases significantly faster. Combined with our empirical finding from the full-scale \#MeToo study that our method scales near-constantly with the number of topics, this indicates that it is feasible to run both the batched and online versions of our method on even larger data. We include the plots for the remaining numbers of topics in Figure \ref{fig:GPUabstudy-fit-mosaic} in the appendix.

\subsubsection{CPU Ablation Study} To demonstrate the benefits of our GPU implementation, we perform an ablation study comparing CPU and GPU TLDA runtime.

As shown in Table \ref{tab:cpuMetoo} (which does not include pre-processing time), we find that the CPU version of our method has relatively constant runtime with respect to the number of topics and takes under 1 hour to converge for each number of topics on the full \#MeToo dataset. Furthermore, we perform an empirical scaling study (see Figure \ref{fig:CPUGPUabstudy-fit-mosaic} in the appendix), in which we confirm that the convergence time of our method scales linearly with the number of topics on CPU as well as on GPU. These results indicate that our method is feasible to use on CPU for datasets on the scale of tens of millions of documents for researchers who do not have access to GPU machines. However, the GPU implementation results in significant gains in TLDA convergence speed and allows for scaling to hundreds of millions and billions of documents. 

\begin{table}[!t]
\centering
\caption{Comparison of CPU and GPU Runtime on the \#MeToo Dataset (7.97 Million Tweets)}\vspace{-8pt}
\label{tab:cpuMetoo}
\resizebox{0.8\linewidth}{!}{
\begin{tabular}{cccc}
\toprule
\textbf{\begin{tabular}[c]{@{}c@{}}Number of \\ Topics (K)\end{tabular}} & \textbf{\begin{tabular}[c]{@{}c@{}}Online TLDA\\ Time on CPU (s)\end{tabular}} & \textbf{\begin{tabular}[c]{@{}c@{}}Online TLDA\\ Time on GPU (s)\end{tabular}} & \textbf{Speedup}     \\
                                           \midrule
10                                         & 1958           & 103            & 19$\times$ %
\\
20                                         & 1906           & 96             & 20$\times$ %
\\
40                                         & 1937           & 98             & 20$\times$ %
\\
60                                         & 1975          & 126            & 16$\times$ %
\\
80                                         & 1947           & 120           & 16$\times$ %
\\
100                                        & 1958         & 103            & 19$\times$ %
\\
\bottomrule
\end{tabular}
}
\end{table}%

\subsection{Qualitative Analysis of the \#MeToo Movement}\label{ssec:qualitative}
Finally, we present a full-scale analysis of the evolution of two years of \#MeToo Twitter discussion over time.  Previous studies have shown the importance of accounting for the dynamic nature of conversation in the \#MeToo movement \citep{liu2019finding}, and here we analyze the topical development of tweets concerning \#MeToo  from just before the start of the movement in August 2017 through September 2019. Understanding the topical development of the \#MeToo movement would be important for scholars who wish to test theories about how social movements organize online, for studies of feminism or gender politics, for two examples.  

To develop this dynamic analysis,  we iteratively grow the corpus of tweets and estimate the TLDA model. That is, first we fit the model on August and September 2017 data to capture the discussion immediately preceding the time when \#MeToo discussion went viral on Twitter. Then we add the next month and estimate the entire model again. We repeat this process for each subsequent month until we reach the end of the dataset. In Figure \ref{fig:dyn_topic_evo}, we report the topic for selected months with the largest weight $\alpha_i$ for both pro- and counter-\#MeToo topics. We call these topics the most prominent. The relative size of the labels indicates the relative prominence of the respective topics. In Figure \ref{fig:dyn_topic_pol_evo}, we report the political topic with the largest $\alpha_i$, which we call the most politically salient.  

This approach leads to three key qualitative findings: first, topics related to politically salient news events are generally ephemeral, changing often as we grow the dataset over time.  In contrast, topical prominence related to personal testimonies, coordinating protests, and  supporting other participants in the \#MeToo movement is persistent over time. Third, discussion around counter-\#MeToo topics declines in prominence over time and is subsumed into one topic by September 2019. These findings suggest avenues for the study of the temporal evolution of political and social coordination  in mass movements, especially on social media, as most research using this type of data to study social movements has not examined the longer-term dynamics of these forms of political engagement \citep{steinert-threlkeld2017,jost_etal_2018}.  Moreover, these results indicate that dynamic changes in topic evolution on social media may be in response to changing news events, another area ripe for new research using large social media and news media datasets using methods like TLDA.

\subsection{Election 2020: Elite Political Communication and the Losers' Effect}

In this second application, we look at another source of real-time social media data relevant to how political elites communicate with and coordinate their supporters, especially in the face of electoral defeat. A peaceful transfer of power is a necessary indicator of a healthy democracy. Studies into how political elites respond to electoral defeat offer critical insights into how democracies persist, even in polarized or politically fraught periods of history.  In light of the 2020 election, longstanding questions around how politicians organize their supporters around election were especially salient. Previous studies into political anger and the losers' effect have relied on survey-based evidence, and produced valuable insights into voter psychology and political behavior \cite{sinclair2018, Craig_Martinez_Gainous_Kane_2006}; however, these methods are static and rely on recalled emotions after an election. Building on this important work, our methodology allows researchers to study a how a politicians coordinates with his or her politically engaged supporters online. With this type of data, we can study online political behaviors in real-time -- President Trump was active online and especially on Twitter \cite{li_2023}, so these conditions allow us to better understand how electoral losers respond to electoral defeat. For policy makers, such methods might allow for real-time detection of the spread of effective attempts to inform the public, while highlighting messaging that angers, rather than informs, the voting public.

As with \#MeToo, this data was collected based on a keyword search employing the social media data collection techniques in \cite{DBLP:journals/corr/abs-2005-02442}  and \cite{li_2023}. The data are lightly structured, so our method can offer insights into the latent structure of this politically engaged demographic: online supporters. In this case, a comprehensive collection of tweets related to keywords related to the administration of the 2020 Presidential election served as the basis of the data collection. Overall, the data is comprised of 29,711,862 tweets collected from September 1, 2020 through the Inauguration on January 20, 2021.  These data were collected in real time as the tweets were being posted. In large part thanks to this real-time collection of data, we were able to capture many tweets prior to their being deleted or moderated, giving us a  unfiltered look at social media activity during a critical period in American electoral politics. 

\begin{figure}
    \centering
    \includegraphics[width=0.75\linewidth]{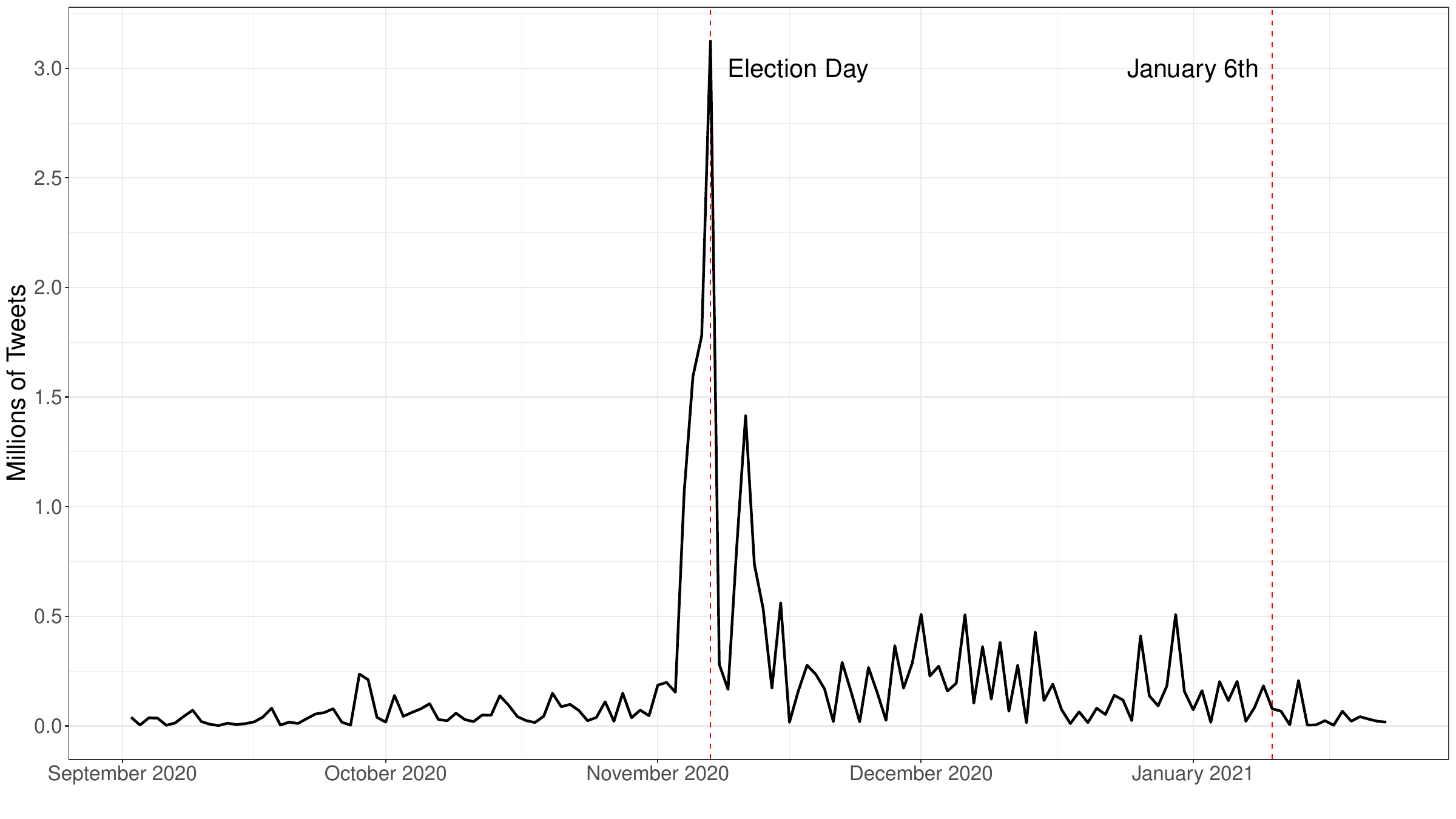}
    \caption{Number of Tweets per Day}
    \label{fig:tweet.per.day}
\end{figure}

In Figure \ref{fig:tweet.per.day}, we report the daily number of tweets collected by our keyword search. The data are most voluminous on November 4th, the day of the election, with nearly 3 million election-related tweets. There is a second peak of just under 1.5 million tweets on November 7th, the day the media outlets declared Joe Biden the winner of the election. Activity remains high following the election with nearly 250 thousand to 500 thousand tweets following the election.

 The TLDA framework offers potential new insights as a data discovery method for a critically important area of political science -- how the public and politicians organize messaging surrounding the legitimacy of elections on social media. Large scale social media data concerning the legitimacy of the election are particularly suited for the TLDA framework because discerning meaningful structure from these data would be impractical without automated methods. More than merely large, the data are unstructured because they are directly collected in real-time from people's authentic online thoughts. In terms of generating new theories about the role of online behavior in disseminating information about the election amplifying the loser's effect, actively organizing political supporters, and its direct role in helping to coordinate the rally and riots on January 6th, a data discovery method like TLDA is the most computationally feasible way to study these data real-time and at scale. 

\subsection{Qualitative  Analysis of Trump-Related Social Media Activity in the 2020 Election}

 We now report an analysis based on the outputs of our TLDA analysis -- we hope this serves as an example of how political scientists might employ our framework to engage in data discovery for unstructured data at large scale. We report descriptive findings on a full population of tweets covering a topic of critical importance to students of political behavior and electoral politics. To that end, we recover 30 topics overall. For this analysis, we classify the topics into 3 main categories based on the following criteria: of the discovered topics, we find three broadly related categories: discrediting the election, discussing legal challenges to the election, and tweets verifying the election result.

\begin{figure}
    \centering
    \includegraphics[width=0.75\linewidth]{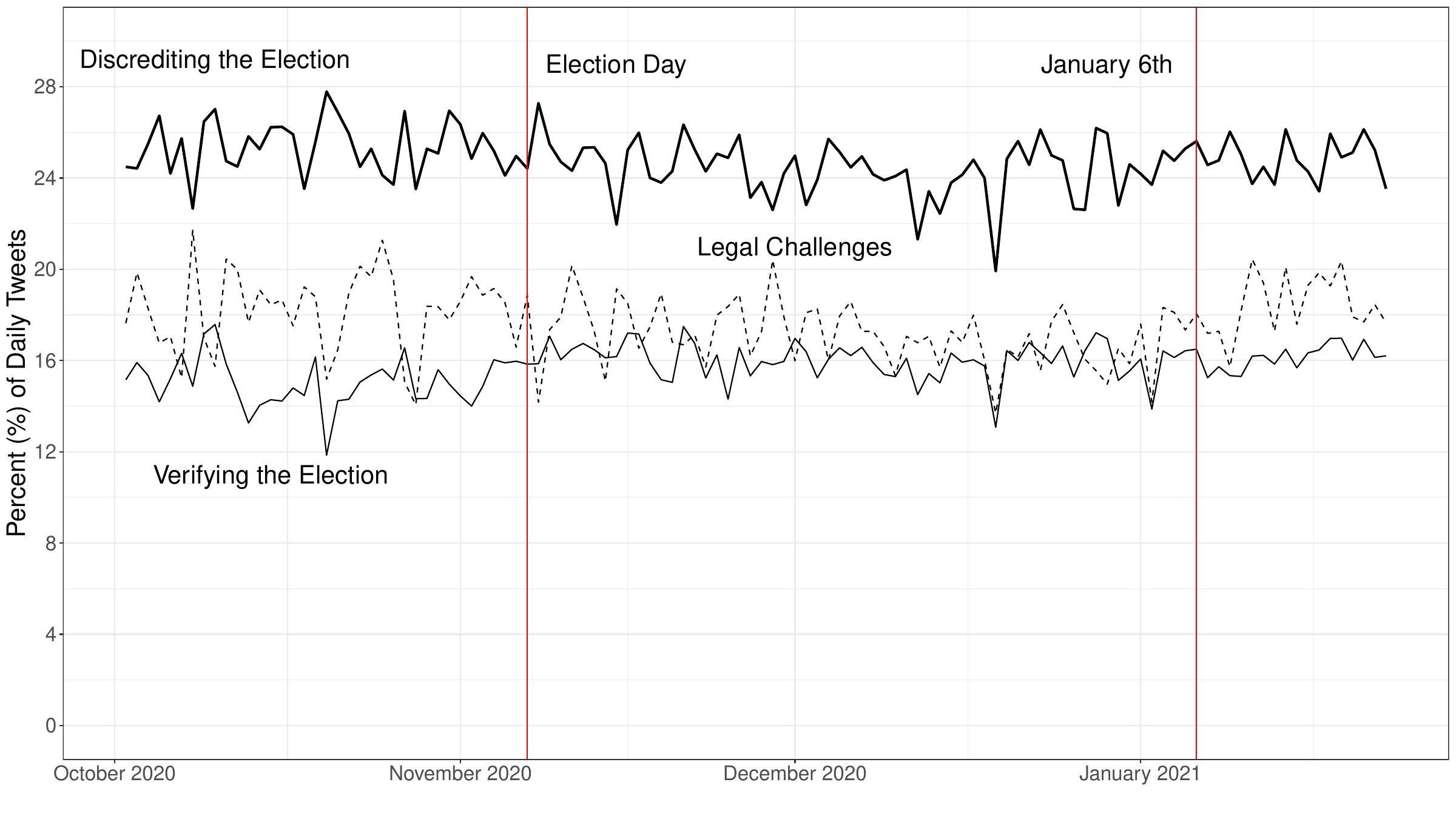}
    \caption{Topical Composition over Time: In this figure, we report the average share of tweets belonging to one of three main categories of topics with a greater than 90 percent probability. }
    \label{fig:topic.comp}
\end{figure}

Notable, in Figure \ref{fig:topic.comp}, we observe relative stability in the daily share of tweets belonging to each category both across time and between the relative share of each of the categories. We see that tweets discrediting the election account for 25 percent of all tweets on average, with a relative peak on January 6th. We see than in the days after January 6th, the share of tweets related to verifying the election results actually drops.

\subsubsection{Pre-processing} Following the same pre-processing as for the \#MeToo data (i.e. we removed all words that appear in fewer than 0.002 percent of documents and in more than 50 percent of documents), we generate a vocabulary of 3788 words. As in the case of the \#MeToo analysis, we perform all pre-processing using the Rapids GPU Framework. We note that preprocessing the entire dataset takes only 3,661 seconds (or 1.02 hours).

\subsubsection{Timing}

In addition, to demonstrate that our method can scale to billions of documents, we use a large COVID dataset to simulate a dataset of $1,043,009,932$ documents. To do so, we loop through the entirety of the data 4 times and update the online version of the method on each batch once within each iteration. We report full results in the Appendix.

In Table~\ref{tab:tlda_time_EF}, we show the model converges in 3.47 hours. We include the top 20 words from each topic and word clouds for the top 3 topics in Section \ref{ssec:topWords} in the appendix to illustrate that our model produces coherent topics. We also find that our method scales linearly to the simulated data, as it takes 13.16 hours to analyze 1.04 billion documents. This illustrates the feasibility of using our method for tuning, estimating, and analyzing topic models on social media data at a large scale.

\section{Conclusion}
Applied researchers have access to larger and larger datasets of text information. 
These datasets may come from social media platforms like Twitter, Reddit, or Facebook~\citep{steinert-threlkeld2018}, congressional speeches, news media archives, legislative text, campaign websites, and senatorial memos. These data sources are being used for critical new studies in political science, across a variety of topics, including political behavior, political opinion, protest movements, legislative agendas, and collective action.  Other large and complex text datasets are being compiled and used in the humanities \citep{dennis-henderson}.  
The widespread availability of unique and high-dimensional text data is opening new doors for applied researchers in the field.

However, off-the-shelf methods such as LDA are computationally inefficient and unusable for analyzing large, high-dimensional text datasets. Our approach provides feasible scale and a method with theoretical foundations for its statistical properties. This compares favorably, as we noted earlier, with proprietary LLMs. New LLM methods show increasing promise, but yet pose lingering challenges for academic researchers, including a lack of strong statistical theoretical foundations.  We hope in future research to show that these statistical properties serve as a foundation for further validation of the quantities of  interest researchers aim to study.

In this paper, we proposed an efficient and theoretically-founded TLDA approach for estimating topic models, which we have shown can be fitted on large text datasets many orders of magnitude faster than existing off-the-shelf topic modeling methods. Our implementation of the TLDA method builds on the TensorLy library. We aim to democratize the analysis of large-text datasets by offering a method that scales on both GPU and CPU, is fully open-source, and whose mathematical underpinnings are clearly communicated.   We hope the wide availability of this method will enable new lines of research on previously out-of-reach scales.

\section{Acknowledgements}
The authors thank Ryan P. Kennedy and the participants at the 2024 ''TexMeth'' Workshop for their comments and suggestions.  Some of SK's work on this project was supported by Summer Undergraduate Research Fellowships from Caltech. We thank the Anima AI + Science Lab and the Alvarez Lab for their comments and feedback.

\section{Author Contributions}
Conceptualization:  SK, DE, JK, AL, RMA, AA; Data collection and curation: SK, DE, RMA; Formal analysis:  SK, DE, JK, AL, AA; Funding acquisition:  RMA and AA; Methodology:  SK, DE, JK, AL, RMA, AA; Project administration:  RMA and AA; Software:  SK, DE, JK; Writing and editing:  SK, DE, JK, AL, RMA, AA.

\section{Data Availability Statement}
Replication code for this article is available via Dataverse at ~\cite{ebanks2025data} at \url{https://doi.org/10.7910/DVN/OKPRJG}.

\bibliographystyle{apsr}
\bibliography{bibliography}

\clearpage

\setcounter{section}{1}
\setcounter{page}{1}
\section*{Online Appendix}
\setcounter{table}{0}
\renewcommand{\thetable}{A\arabic{table}}
\setcounter{figure}{0}
\renewcommand{\thefigure}{A\arabic{figure}}

In this appendix, we first collect the notation used throughout the paper (Sec.~\ref{ssec:notation}), detail the general form of the cumulants (Sec.~\ref{ssec:uncentered-cumulants}), and provide additional information on the hardware used for the experiments, the exact hyper-parameters used. We also show additional results, including topics recovered with our method and ablation studies on both real and synthetic data. 

\subsection{Summary of notation and mathematical background}\label{ssec:notation}

Innovations in spectral decomposition methods from low-order tensors have allowed for increasingly parsimonious estimation of latent variable models. To connect to previous work in this area, we use notation consistent with~\cite{JMLR:v15:anandkumar14b}. A real-valued $p$-th order tensor $\tensor{A} \in \bigotimes^p_{i=1} \R^{n_i}$ is a member of the tensor product of Euclidean spaces $\R^{n_i}, i \in [p]$. In our case, we have that all $n_i$ are equal due to the nature of the model considered, and so we denote the product simply as $\tensor{A} \in \bigotimes^p_{i=1} \R^n$.
For a vector $\vector{v} \in \R^n $, we use
$\vector{v}{\otimes^p} 
:= \vector{v} \otimes \vector{v} \otimes \dots \otimes \vector{v} \in \bigotimes^p \R^n$ to denote its $p$-th tensor power. 
In this work, we use tensors of order at most $3$. For such tensors, note that as with vectors (where $p = 1$) and matrices (where $p = 2$), we may identify a
$p$-th order tensor with the $p$-way array of real numbers $[A_{i_1,i_2,...,i_p}
: i_1, i_2, . . . , i_p \in [n]]$, where
$A_{i_1,i_2,...,i_p}$
is the $(i_1, i_2, . . . , i_p)$-th coordinate of $A$. Note that the tensor objects considered in this paper are symmetric, meaning that their $p$-way array representations are permutation invariant: that is, for all indices $i_1, i_2, \cdots, i_p \in [n]$, $A_{i_1, i_2, \cdots, i_p} =
A_{i\pi(1), i\pi(2), \cdots ,i\pi(p)}$
for any permutation $\pi$ on $[p]$. This property importantly holds for the 3rd-order moment of the LDA model~\cite{Anandkumar2012}. See~\cite{janzamin2019spectral} for an overview of spectral learning on tensors.

In Table~\ref{tab:notation}, we summarize the notation used throughout the paper.
\begin{table}[htp]
    \caption{Table of Notations used in this paper.}
    \centering
    \resizebox{\columnwidth}{!}{%
    \begin{tabular}{c l c}
    \toprule
    \textbf{Symbol} & \textbf{Meaning} & \textbf{Domain}\\ 
    \midrule
    $K$ & Number of topics & \N
    \\
    $\vector{h}$ & Topic mixture & $\mathbb{R}^K$
    \\

    $V$ & Vocabulary size & \N
    \\
    $\vector{\mu} $ & $\expectation{}{f_i| \vector{h}} = \vector{\mu} \vector{h}$ & $\mathbb{R}^V$
    \\
    $\vector{f}_i $ & Frequency vector for the $i$-th document
    & $\mathbb{R}^V$
    \\
    $\tilde{\vector{x}}_i $ & Centered frequency vector for the $i$-th document & $\mathbb{R}^V$
    \\
    $ \vector{x}_i $  & Centered \& whitened frequency vector & $\mathbb{R}^V$
    \\
    $N$ & Number of documents & \N
    \\
    $D$ & Whitening dimension size & \N
    \\
    $n_b$ & Number of documents in a mini-batch & \N
    \\
    $\matrix{X}$ & centered, whitened matrix with columns $\vector{x}_i$ & $\R^{n_b \times D}$
    \\
    $\Phi$ & learned factors of the decomposition & $\R^{D \times K}$
    \\
    \bottomrule
    \end{tabular}
    }
    \label{tab:notation}
\end{table}

\subsection{Uncentered cumulants}\label{ssec:uncentered-cumulants}
Here, we detail the general form of the first, second, and third-order cumulants used for learning the tensor LDA, from uncentered data.

First, we define the first-order cumulant:
\begin{eqnarray*}
\label{M1}
\vector{M}_1 &=& \frac{1}{N} \cdot \sum_{i=1}^{N} \tensor{f}_i 
\end{eqnarray*}

From this, we define the second-order cumulant:
\begin{eqnarray*}
\label{M2}
\matrix{M}_2 &=& \frac{\alpha_0 + 1}{N} \cdot \sum_{i=1}^{N} \biggl ( (\tensor{f}_i \otimes \tensor{f}_i) - \diag(\tensor{f}_i) \biggr ) - \alpha_0(\matrix{M}_1 \otimes \matrix{M}_1)
\end{eqnarray*}

Finally, we introduce the third-order cumulant:
\begin{eqnarray*}
\label{M3}
\tensor{M}_3 &=& \frac{(\alpha_0 + 1)(\alpha_0 + 2)}{2N} \sum_{i=1}^{N} \biggl[ (\tensor{f}_i \otimes \tensor{f}_i \otimes \tensor{f}_i)\\
&-& (\diag(\tensor{f}_i) \otimes \tensor{f}_i)
- (\tensor{f}_i \otimes \diag(\tensor{f}_i)) \\
&-& \sum_{m=1}^V\sum_{n=1}^V (\tensor{f}_{i,m} * \tensor{f}_{i,n})(\tensor{e}_m \otimes \tensor{e}_n \otimes \tensor{e}_m) \\ 
 &+& 2 \sum_{m=1}^{V} \tensor{f}_{i, m} (\tensor{e}_m \otimes \tensor{e}_m \otimes \tensor{e}_m) \biggr ]\\
&-& \frac{\alpha_0(\alpha_0 + 1)}{2N}\sum_{i=1}^{N} \biggl ( \sum_{m=1}^V \tensor{f}_{i, m}(\tensor{e}_m \otimes \tensor{e}_m \otimes \vector{M}_1)\\
&+& \sum_{m=1}^V \tensor{f}_{i, m}(\tensor{e}_m \otimes \vector{M}_1 \otimes \tensor{e}_m) \\
&+& \sum_{m=1}^V \tensor{f}_{i, m}(\matrix{M}_1 \otimes \tensor{e}_m \otimes \tensor{e}_m) \biggr ) + \alpha_0^2(\vector{M}_1 \otimes \vector{M}_1 \otimes \vector{M}_1)
\end{eqnarray*}

\subsection{Proof of Theorem 1}\label{ssec:proof}
Here, we give the main steps to prove the Theorem~\ref{thm1}.
\begin{proof}
We begin by applying the linearity of the Kronecker product,
\begin{align}\nonumber
\sum_{i = 1}^K \frac{\alpha_i}{\alpha_0} (\vector{\nu}_i + \vector{M}_1) \otimes (\vector{\nu}_i + \vector{M}_1) \otimes (\vector{\nu}_i + \vector{M}_1)\\
\nonumber
= \sum_{i = 1}^K \frac{\alpha_i}{\alpha_0} [\vector{\nu}_i\otimes\vector{\nu}_i\otimes\vector{\nu}_i + \\ \nonumber
(\vector{\nu}_i \otimes \vector{M}_1 \otimes \vector{M}_1 +\vector{M}_1 \otimes \vector{\nu}_i \otimes \vector{M}_1 + \vector{M}_1 \otimes \vector{M}_1 \otimes \vector{\nu}_i)
\end{align}
$$+ (\vector{\nu}_i \otimes \vector{\nu}_i \otimes \vector{M}_1 + \vector{\nu}_i \otimes \vector{M}_1 \otimes \vector{\nu}_i + \vector{M}_1 \otimes \vector{\nu}_i \otimes \vector{\nu}_i) + \vector{M}_1 \otimes \vector{M}_1 \otimes \vector{M}_1 ]$$

We consider the first term, which we call $T_1$. We know that if $\vector{\nu}_i$ are the learned factors, then 

$$T_1 = \sum_{i = 1}^K \frac{\alpha_i}{\alpha_0} \vector{\nu}_i\otimes\vector{\nu}_i\otimes\vector{\nu}_i = \mathbb{E}[\tilde{\tensor{{M}}_3}]$$ 

$$= \mathbb{E}\left [\frac{(\alpha_0 + 1)(\alpha_0 + 2)}{2} (\vector{f}_1 - \vector{M}_1)\otimes (\vector{f}_2 - \vector{M}_1) \otimes (\vector{f}_3 - \vector{M}_1)\right]$$

Here, $\vector{f}_1, \vector{f}_2, \vector{f}_3$ are variables representing the first, second, and third document for which the probability of co-occurrence is computed. Thus, they are interchangeable in terms with $3$ or fewer variables. Using this, linearity of expectation, and the equality $\mathbb{E}[\vector{M}_1] = \vector{M}_1$, if we denote the first term to be $T_1$ we can show that
$$T_1 = \frac{(\alpha_0 + 1)(\alpha_0 + 2)}{2}{\big(} [\mathbb{E}[\vector{f}_1 \otimes \vector{f}_2 \otimes \vector{f}_3] - \mathbb{E}[\vector{f}_1 \otimes \vector{f}_2 \otimes \vector{M}_1]$$ $$- \mathbb{E}[\vector{f}_1 \otimes \vector{M}_1 \otimes \vector{f}_2] - \mathbb{E}[\vector{M}_1 \otimes \vector{f}_2 \otimes \vector{f}_2] + 2(\vector{M}_1 \otimes \vector{M}_1 \otimes \vector{M}_1) {\big)}$$

Next, we consider the terms in the original equation with $1$ instance of $\vector{\nu}_i$. There, we note that $\sum_{i = 1}^K \frac{\alpha_i}{\alpha_0} \vector{\nu}_i = \mathbb{E}[\tilde{\vector{M}}_1] = \vector{0}$ and use the linearity of the Kronecker product to simplify each of these terms to $\vector{0}$.

Similarly, we consider the terms in the original equation with $2$ instances of $\vector{\nu}$. First, we note that as our data is centered,
$$\sum_{i = 1}^K \frac{\alpha_i}{\alpha_0} \vector{\nu}_i \otimes \vector{\nu}_i = \mathbb{E}[\tilde{\vector{M}}_2] = (\alpha_0 + 1)\mathbb{E}[(\vector{f}_1 - \vector{M}_1) \otimes (\vector{f}_2 - \vector{M}_1)] $$ $$= (\alpha_0 + 1)\mathbb{E}[\vector{f}_1 \otimes \vector{f}_2 - \vector{M}_1 \otimes \vector{f}_2 - \vector{f}_1 \otimes \vector{M}_1 + \vector{M}_1 \otimes \vector{M}_1]$$
$$= (\alpha_0 + 1)(\mathbb{E}[\vector{f}_1 \otimes \vector{f}_2] - \vector{M}_1 \otimes \vector{M}_1)$$

Now, we denote the sum of the terms with $2$ $\vector{\nu}$ values to be $T_2$. We once again apply the linearity of the Kronecker product to obtain
$$T_2 = (\alpha_0 + 1)(\mathbb{E}[\vector{f}_1 \otimes \vector{f}_2] - \vector{M}_1 \otimes \vector{M}_1) \otimes \vector{M}_1$$ $$+ (\alpha_0 + 1)(\mathbb{E}[\vector{f}_1 \otimes \vector{M}_1 \otimes \vector{f}_2] - \vector{M}_1 \otimes \vector{M}_1 \otimes \vector{M}_1) $$ $$+ \vector{M}_1 \otimes (\alpha_0 + 1)(\mathbb{E}[\vector{f}_1 \otimes \vector{f}_2] - \vector{M}_1 \otimes \vector{M}_1)$$
By the Kronecker product's linearity and the equality $\mathbb{E}[\vector{M}_1] = \vector{M}_1$,

$$T_2 = (\alpha_0 + 1)\big{ [}\mathbb{E}[\vector{f}_1 \otimes \vector{f}_2 \otimes \vector{M}_1] + \mathbb{E}[\vector{f}_1 \otimes \vector{M}_1 \otimes \vector{f}_2] + \mathbb{E}[\vector{M}_1 \otimes \vector{f}_1 \otimes \vector{f}_2]$$ $$- 3(\vector{M}_1 \otimes \vector{M}_1 \otimes \vector{M}_1)\big{]}$$

Finally, we compute the entire sum. As $\sum_{i = 1}^K\frac{\alpha_i}{\alpha_0} = 1$, we can move the $\vector{M}_1\otimes ^3$ term out of the summation and simplify:
$$\sum_{i = 1}^K \frac{\alpha_i}{\alpha_0} (\vector{\nu}_i + \vector{M}_1) \otimes (\vector{\nu}_i + \vector{M}_1) \otimes (\vector{\nu}_i + \vector{M}_1) = T_1 + T_2 + \vector{M}_1 \otimes \vector{M}_1 \otimes \vector{M}_1$$ 
$$= \frac{(\alpha_0 + 1)(\alpha_0 + 2)}{2}\mathbb{E}[\vector{f}_1 \otimes \vector{f}_2 \otimes \vector{f}_3]$$ $$- \frac{\alpha_0(\alpha_0 + 1)}{2}\left( \mathbb{E}[\vector{f}_1 \otimes \vector{f}_2 \otimes \vector{M}_1] + \mathbb{E}[\vector{f}_1 \otimes \vector{M}_1 \otimes \vector{f}_2] + \mathbb{E}[\vector{M}_1 \otimes \vector{f}_1 \otimes \vector{f}_2] \right )$$ $$+ \alpha_0^2(\vector{M}_1 \otimes \vector{M}_1 \otimes \vector{M}_1) = \mathbb{E}[\tensor{M}_3]$$
So $\sum_{i = 1}^K \frac{\alpha_i}{\alpha_0} (\vector{\nu}_i + \vector{M}_1) \otimes (\vector{\nu}_i + \vector{M}_1) \otimes (\vector{\nu}_i + \vector{M}_1) = \mathbb{E}[\tensor{M}_3]$ as desired.
\end{proof}

\begin{figure*}[ht]
    \begin{subfigure}[b]{.32\linewidth}
        \includegraphics[width=\linewidth]{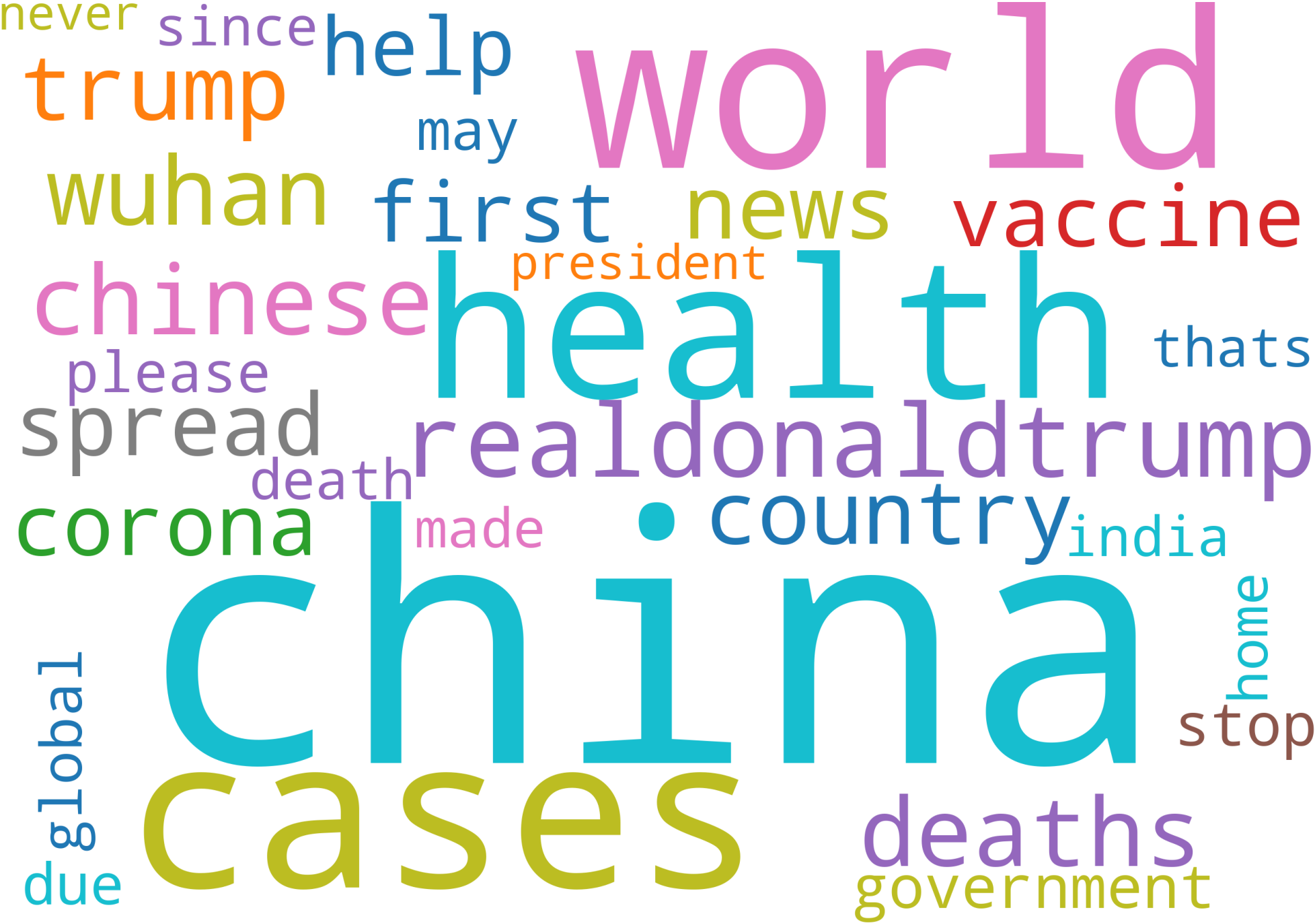}
        \caption{\textbf{Topic 0}}
    \end{subfigure}
    \hfill
    \begin{subfigure}[b]{.32\linewidth}
        \includegraphics[width=\linewidth]{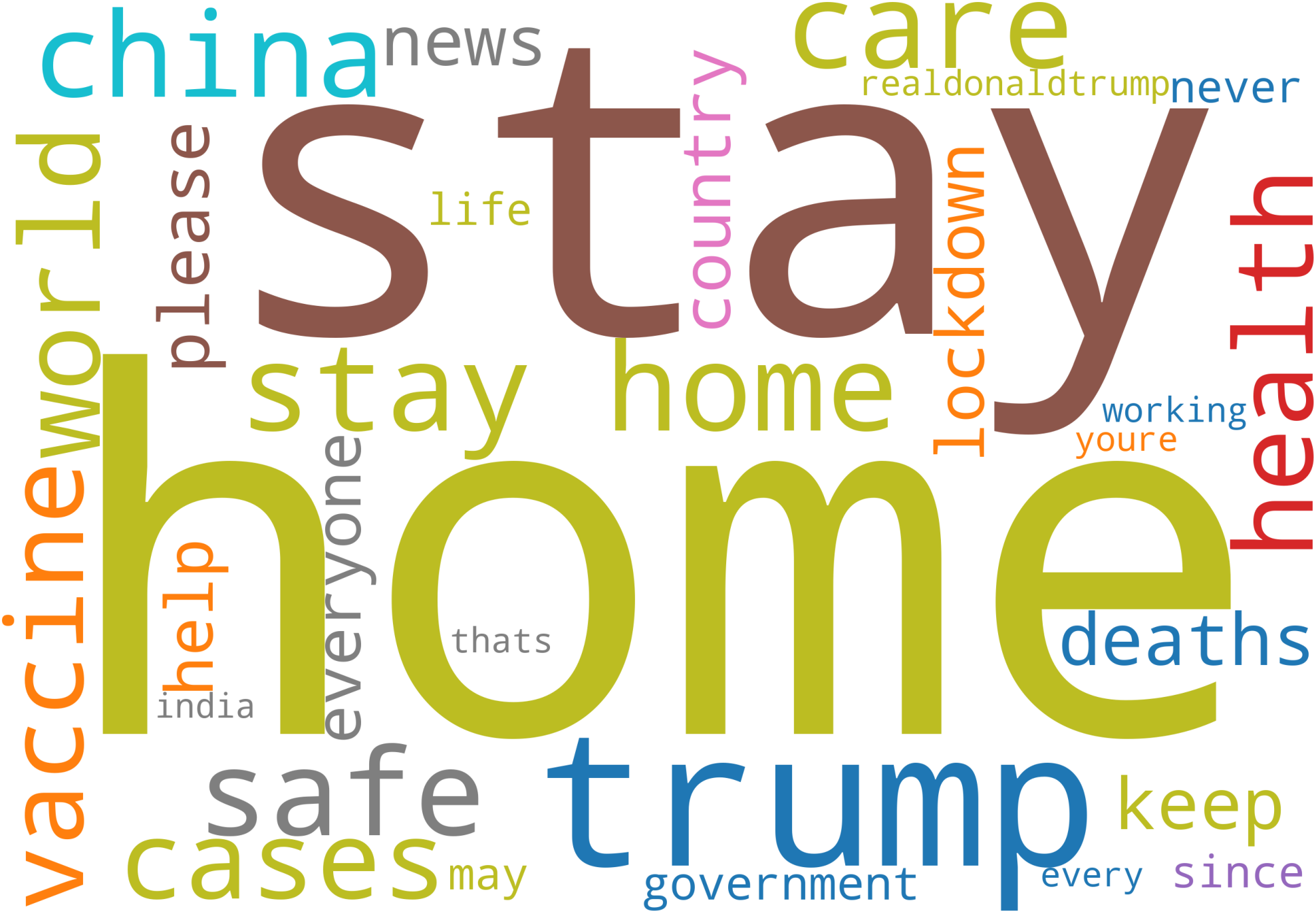}
        \caption{\textbf{Topic 1}}
    \end{subfigure}
    \hfill
    \begin{subfigure}[b]{.32\linewidth}
        \includegraphics[width=\linewidth]{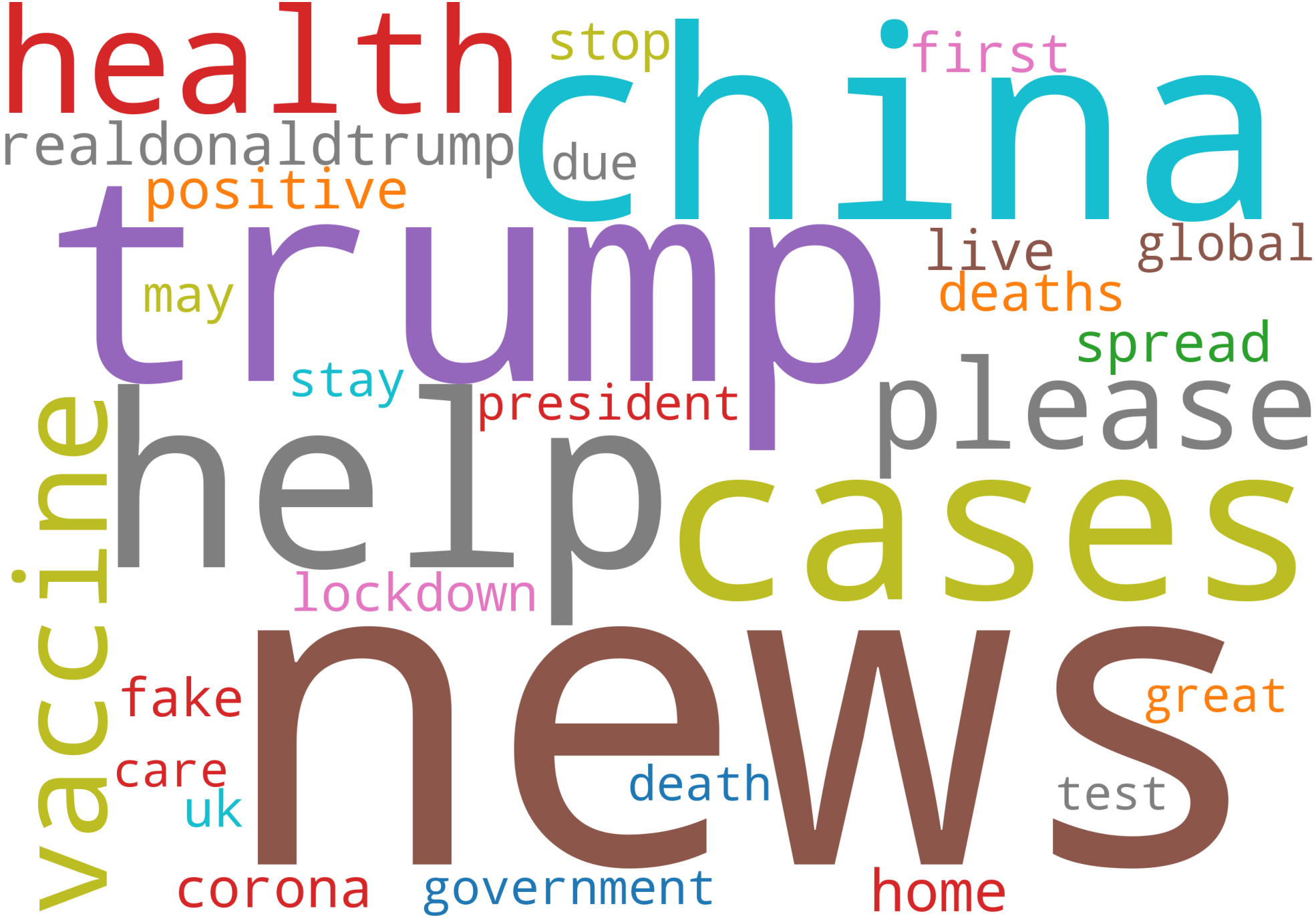}
        \caption{\textbf{Topic 3}}
    \end{subfigure}
    \caption{Wordclouds for COVID-19 Dataset}
    \label{fig:wordclouds}
\end{figure*}

\subsection{Simulation Data Generation Process}
\label{sec:simDGP}

The provided Python code simulates a Latent Dirichlet Allocation (LDA) model. Here's a summary of the data generation process in LaTeX math and words:

Constants are defined for the number of topics $K$, vocabulary size $V$, number of documents $N$, and terms per document $f_i$.

Dirichlet priors are set for the topic-word distribution ($\beta$) and the document-topic distribution ($\alpha$). Intuitively:

\begin{itemize}
    \item 

For each topic, a multinomial distribution over words is generated ($\mu$), which represents the probability of each word given a topic.

For each document, a multinomial distribution over topics is generated ($\theta$), which represents the probability of each topic given a document.

\end{itemize}

Written out fully:

\begin{itemize}
    \item  For each term in a document, a topic ($h$) is sampled from the document's topic distribution ($\theta$), and a word ($F_i$) is sampled from the topic's word distribution (mu).

\item For each topic $k \in {1, ..., K}$, draw a distribution over words $\mu_k \sim \text{Dirichlet}(\beta)$.

\item For each document $d \in {1, ..., D}$, draw a distribution over topics $\theta_d \sim \text{Dirichlet}(\alpha)$.

\item For each word $n \in {1, ..., N}$ in each document $d$, draw a topic assignment $z_{dn} \sim \text{Multinomial}(\theta_d)$ and a word $w_{dn} \sim \text{Multinomial}(\mu_{z_{dn}})$.

\end{itemize}

We then take this simulated data and then fit
a TLDA model to the generated data and tests the accuracy of the model by comparing the learned topic-word distributions to the true distributions. The accuracy is measured by the correlation between the true and learned distributions.

\subsection{Top Words Produced by TLDA}\label{ssec:topWords}

In this section, we report the top 20 words for each topic produced by the batched TLDA model when run with $K = 10$ topics for the full \#MeToo dataset in Table \ref{tab:metootopics} and the top 20 words for each topic produced by the online TLDA model when run with $K = 5$ topics for the full COVID dataset in Table \ref{tab:covidtopics}. We manually select the topic labels by summarizing the top words.

We also provide word clouds for the COVID-19 topics with the top three highest weights $\alpha_i$ in Figure \ref{fig:wordclouds}. These topics relate to the topics of China and COVID, staying safe during COVID, and news and reporting on COVID, respectively.

\begin{table}[!ht]
\centering
\caption{Top words in each \#MeToo Topic for $K = 10$}
\label{tab:metootopics}
\resizebox{0.8\linewidth}{!}{%
\begin{tabular}{cll}
\toprule
\textbf{Topic} & \multicolumn{1}{c}{\textbf{Topic Label}}                                                & \multicolumn{1}{c}{\textbf{Top 20 Words}}                                                                                                                                                                                  \\
\midrule
0              & \begin{tabular}[c]{@{}l@{}}\#MeToo politics and \\ anti-Trump\end{tabular}            & \begin{tabular}[c]{@{}l@{}}vote, support, kavanaugh, rape, \#believesurvivor, \\ stori,  victim, everi, believ, march, senat, \\ accus, love, power, \#imwithh, pleas, never, \\ new, democrat, @realdonaldtrump\end{tabular}                \\
1              & \begin{tabular}[c]{@{}l@{}}Brett Kavanaugh and \\ Dr. Ford hearings\end{tabular} & \begin{tabular}[c]{@{}l@{}}\#believesurvivor, ford, kavanaugh, believ, dr, \\ support,  survivor, accus, stand, victim, stori, \\ senat, rape, blasey, blasey ford, alleg,  countri, \\christin, \#believewomen, march\end{tabular}                         \\
2              & Support victims    & \begin{tabular}[c]{@{}l@{}}rape, victim, accus, girl, countri, support,\\ believ, stori, alleg, march, stop, survivor, \\sex, report, \#whyididntreport, never, love, \\ \#believesurvivor, speak, new\end{tabular}                  \\
3              &  \begin{tabular}[c]{@{}l@{}}Taking action in support \\ of \#MeToo\end{tabular}              & \begin{tabular}[c]{@{}l@{}}sign, pleas, support, rape, victim, believ,\\ accus, stori, march, chang, love, power, \\share, new, stand, girl, never, help,\\ everi, best\end{tabular} \\
4              & Reporting allegations                                                                          & \begin{tabular}[c]{@{}l@{}}report, tweet, rape, \#whyididntreport, victim, \\ support, stori, believ, accus, never, new, \\ march, power, love, alleg, speak, thought, \\ pleas, pm, survivor\end{tabular}              \\
5              & \begin{tabular}[c]{@{}l@{}}Protesting on behalf of \\ victims around the world\end{tabular}                                                                       & \begin{tabular}[c]{@{}l@{}}protest, rape, believ, support, march, stori, \\accus, victim, love, realli, world, speak, \\around, new, power, never, stand, \#believesurvivor, \\feminist, last\end{tabular}          \\
6              & \begin{tabular}[c]{@{}l@{}}Supporting victims\\ and \#heforshe\end{tabular} & \begin{tabular}[c]{@{}l@{}}great, support, stori, rape, accus, march, \\love, victim, speak, believ, realli, new,\\ help, power, never, stand, girl, \#heforsh,\\ start, world\end{tabular}                           \\
7              & Believe survivors      & \begin{tabular}[c]{@{}l@{}}believ, speak, rape, accus, support, stori,\\ victim, \#believesurvivor, survivor, ford, \\ \#whyididntreport, much,  march, lie, new, equal,\\ never, love,  stand, realli\end{tabular}            \\
8              & Help survivors                                                                         & \begin{tabular}[c]{@{}l@{}}survivor, life, \#believesurvivor, believ, support, \\rape, stori, victim, stop, dc, stand,\\ power, love, help, march, everi, speak, \\ new, report, chang\end{tabular}                                      \\
9              & \begin{tabular}[c]{@{}l@{}}Women's March \end{tabular}          & \begin{tabular}[c]{@{}l@{}}victim, rape, support, accus, speak, believ, \\stori, help, million, march, never, everi, stand, \\survivor, love, real, alleg, power, new, lie\end{tabular}\\
\bottomrule
\end{tabular}
}
\end{table}

\begin{table}[!ht]
\centering
\caption{Top words in each COVID-19 Topic for $K = 5$}
\label{tab:covidtopics}
\resizebox{.5\textwidth}{!}{
\begin{tabular}{cll}
\toprule
\textbf{Topic} & \multicolumn{1}{c}{\textbf{Topic Label}}                                       & \multicolumn{1}{c}{\textbf{Top 20 Words}}                                                                                                                                                                 \\
\midrule 
0              & \begin{tabular}[c]{@{}l@{}}China and \\ COVID\end{tabular}              & \begin{tabular}[c]{@{}l@{}}china, world, cases, health, realdonaldtrump, chinese,\\ deaths, wuhan, trump, news, country, spread, corona, \\ vaccine,  first, help, government, stop, may, due\end{tabular}     \\
1              & \begin{tabular}[c]{@{}l@{}}Help stop \\ the spread\end{tabular}                     & \begin{tabular}[c]{@{}l@{}}home, stay, trump, stay home, safe, care, china, \\world, cases, health, vaccine, deaths, keep, please, \\ news, country, lockdown, help, everyone, government\end{tabular} \\
2              & \begin{tabular}[c]{@{}l@{}}Reported \\ positive cases\end{tabular}                & \begin{tabular}[c]{@{}l@{}}cases, trump, deaths, china, health, realdonaldtrump, \\ total, world, news, death, confirmed, country, reported,\\ number, positive, president, state, corona, last, india\end{tabular}                         \\
3              & \begin{tabular}[c]{@{}l@{}}News and \\ reporting on \\ COVID\end{tabular} & \begin{tabular}[c]{@{}l@{}}news, trump, china, help, cases, health, vaccine,\\ please, realdonaldtrump, home, live, spread, positive,\\ corona, first, deaths, stop, fake, lockdown, may\end{tabular} \\
4              & \begin{tabular}[c]{@{}l@{}}US government \\ response\end{tabular}      & \begin{tabular}[c]{@{}l@{}}trump, last, deaths, spread, government, positive,\\ china,  may, vaccine, home,  india, lockdown, due,  \\ health,  help, world, test, since, days, please\end{tabular}\\
\bottomrule
\end{tabular}
}
\end{table}

\subsection{Comparison Against other LDA Scalable Methods}

We note that our method is around 3-4x faster than WarpLDA. Furthermore, unlike WarpLDA, our method has theoretical convergence guarantees, as shown by \cite{anandkumar2013spectral} in prior work on TLDA. Thus, not only is our method faster than prior scalable LDA work, but it also allows for the benefits of  TLDA, such as these convergence guarantees and exploiting third-order word co-occurrences, to be used for modeling large-scale data for the first time.

\begin{table}[!ht]
    \caption{Updated TLDA timing comparison on full \#MeToo dataset}\label{tab:new_comp}
    \centering
    \begin{tabular}{ccc}
    \toprule
    Number of Topics & Model & Convergence Time (s) \\
    \midrule
    10 & Gensim LDAMulticore & 1581.80 \\
     & WarpLDA & 415.95 \\
     & Batched TLDA & 121.81 \\
     & Online TLDA & 103.45 \\
    \midrule
    20 & Gensim LDAMulticore & 1528.38 \\
     & WarpLDA & 348.32 \\
     & Batched TLDA & 145.33 \\
     & Online TLDA & 96.10 \\
    \midrule
   40 & Gensim LDAMulticore & 8801.33 \\
     & WarpLDA & 321.18 \\
     & Batched TLDA & 149.54 \\
     & Online TLDA & 98.27 \\
    \midrule
    60 & Gensim LDAMulticore & 14448.87 \\
     & WarpLDA & 396.65 \\
     & Batched TLDA & 152.67 \\
     & Online TLDA & 126.19 \\
    \midrule
    80 & Gensim LDAMulticore & 14700.59 \\
     & WarpLDA & 406.55 \\
     & Batched TLDA & 158.88 \\
     & Online TLDA & 120.13 \\
    \midrule
    100 & Gensim LDAMulticore & 14830.54 \\
     & WarpLDA & 413.43 \\
     & Batched TLDA & 171.82 \\
     & Online TLDA & 103.45 \\
    \bottomrule
    \end{tabular}
\end{table}

\subsection{Ablation Study Runtime Comparison Plots}

\begin{figure*}
    \centering
    \begin{subfigure}[b]{.48\linewidth}
        \includegraphics[width=\linewidth]{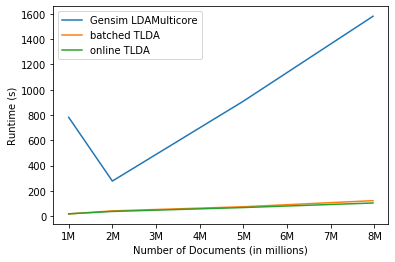}
        \caption{\textbf{10 topics}}
        \label{fig:GPUabstudyfit-total10}
        \vspace{10pt}
    \end{subfigure}
    \hfill
    \begin{subfigure}[b]{.48\linewidth}
        \includegraphics[width=\linewidth]{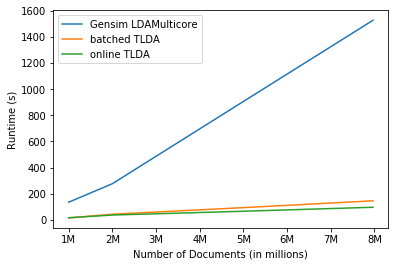}
        \caption{\textbf{20 topics}}
        \label{fig:GPUabstudyfit-total20}
        \vspace{10pt}
    \end{subfigure}
    
    \begin{subfigure}[b]{.48\linewidth}
        \includegraphics[width=\linewidth]{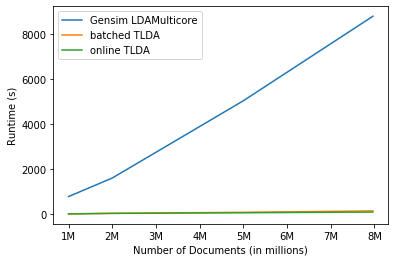}
        \caption{\textbf{40 topics}}
        \label{fig:GPUabstudyfit-total40}
        \vspace{10pt}
    \end{subfigure}
    \hfill
    \begin{subfigure}[b]{.48\linewidth}
        \includegraphics[width=\linewidth]{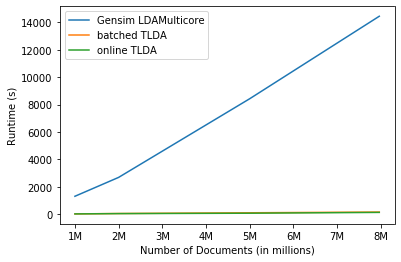}
        \caption{\textbf{60 topics}}
        \label{fig:GPUabstudyfit-total60}
        \vspace{10pt}
    \end{subfigure}

    \begin{subfigure}[b]{.48\linewidth}
        \includegraphics[width=\linewidth]{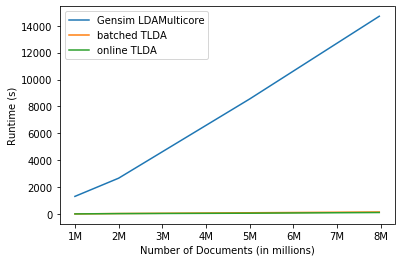}
        \caption{\textbf{80 topics}}
        \label{fig:GPUabstudyfit-total80}
        \vspace{10pt}
    \end{subfigure}
    \hfill
    \begin{subfigure}[b]{.48\linewidth}
        \includegraphics[width=\linewidth]{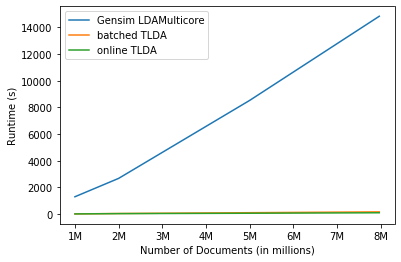}
        \caption{\textbf{100 topics}}
        \label{fig:GPUabstudyfit-total100}
        \vspace{10pt}
    \end{subfigure}

    \caption{\textbf{Runtime comparison for TLDA on GPU vs Gensim} for various number of topics. For each number of topic, we plot the time to convergence as a function of the number of documents. Our method scales linearly while Gensim's convergence time is superlinear with the number of documents. The difference is even more pronounced as we increase the number of topics. We note that these plots do not include pre-processing time.}
    \label{fig:GPUabstudy-fit-mosaic}
\end{figure*}

\begin{figure*}
    \centering
    \begin{subfigure}[b]{.48\linewidth}
        \includegraphics[width=\linewidth]{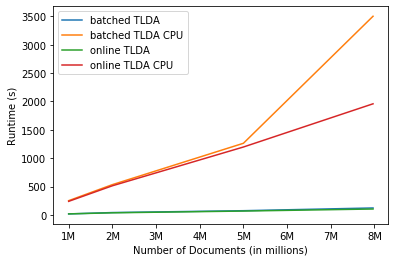}
        \caption{\textbf{10 topics}}
        \label{fig:CPUGPUabstudyfit-total10}
        \vspace{10pt}
    \end{subfigure}
    \hfill
    \begin{subfigure}[b]{.48\linewidth}
        \includegraphics[width=\linewidth]{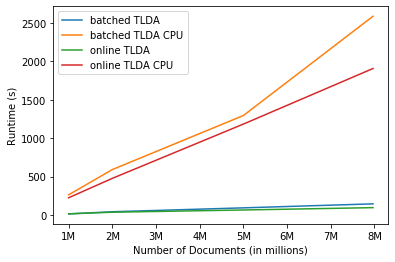}
        \caption{\textbf{20 topics}}
        \label{fig:CPUGPUabstudyfit-total20}
        \vspace{10pt}
    \end{subfigure}
    
    \begin{subfigure}[b]{.48\linewidth}
        \includegraphics[width=\linewidth]{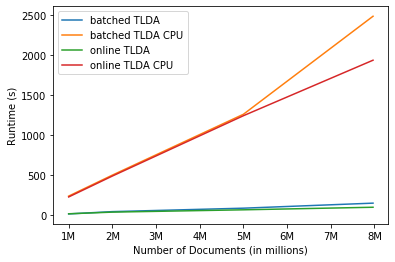}
        \caption{\textbf{40 topics}}
        \label{fig:CPUGPUabstudyfit-total40}
        \vspace{10pt}
    \end{subfigure}
    \hfill
    \begin{subfigure}[b]{.48\linewidth}
        \includegraphics[width=\linewidth]{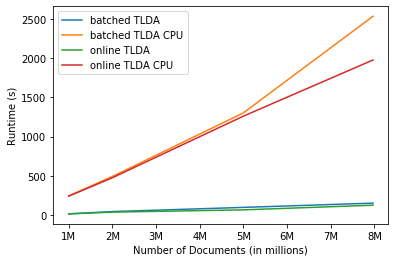}
        \caption{\textbf{60 topics}}
        \label{fig:CPUGPUabstudyfit-total60}
        \vspace{10pt}
    \end{subfigure}

    \begin{subfigure}[b]{.48\linewidth}
        \includegraphics[width=\linewidth]{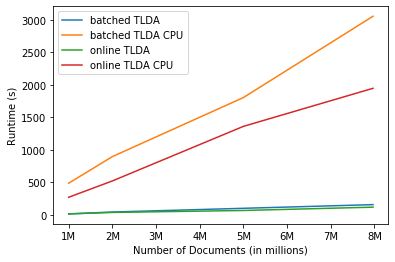}
        \caption{\textbf{80 topics}}
        \label{fig:CPUGPUabstudyfit-total80}
        \vspace{10pt}
    \end{subfigure}
    \hfill
    \begin{subfigure}[b]{.48\linewidth}
        \includegraphics[width=\linewidth]{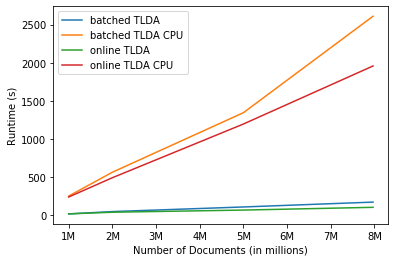}
        \caption{\textbf{100 topics}}
        \label{fig:CPUGPUabstudyfit-total100}
        \vspace{10pt}
    \end{subfigure}

    \caption{\textbf{Runtime comparison for TLDA on CPU vs GPU} for various number of topics. For each number of topic, we plot the time to convergence as a function of the number of documents. Both the CPU and GPU versions of our method scale linearly. We note that these plots do not include pre-processing time.}
    \label{fig:CPUGPUabstudy-fit-mosaic}
\end{figure*}

To illustrate the scalability of our online TLDA approach, compared with Gensim and the batched version, we performed thorough ablations and measured the fitting time for various number of topics (10, 20, 40, 60, 80, and 100 topics), Figure~\ref{fig:GPUabstudy-fit-mosaic}. In each case, we plot the time to convergence as a function of the number of documents, where we vary the number of documents between 1 and 8 million. We also compare our online approach on CPU vs on GPU in Figure ~\ref{fig:CPUGPUabstudy-fit-mosaic}.

\subsection{Experimental Setup}
We list in Table~\ref{tab:machines} the hardware used to run each comparison.  
\begin{table}[h]
\caption{System Used for the Experiments in Each Results Section} 
\label{tab:machines}
\resizebox{\linewidth}{!}{%
\centering
\begin{tabular}{cccc}
\toprule
\textbf{Machine}                                          & \textbf{GPU}                                              & \textbf{CPU}                                                                                     & \textbf{Experiments}                                                      \\
\midrule
\begin{tabular}[c]{@{}c@{}}NVIDIA \\ A100\end{tabular}    & \begin{tabular}[c]{@{}c@{}}A100 SXM4\\ 80GB\end{tabular}  & \begin{tabular}[c]{@{}c@{}}AMD EPYC 7742 \\ 64 cores, 550 GB RAM\end{tabular}                    & \begin{tabular}[c]{@{}c@{}}GPU Experiments \\ in Section 4.1\end{tabular} \\
\begin{tabular}[c]{@{}c@{}}Tesla \\ V100\end{tabular} & \begin{tabular}[c]{@{}c@{}}V100 SXM2 \\ 32GB\end{tabular} & \begin{tabular}[c]{@{}c@{}}Intel(R) Xeon(R) \\ CPU @ 2.20GHz\\ 80 cores, 503 GB RAM\end{tabular} & \begin{tabular}[c]{@{}c@{}}CPU Experiments \\ in Section 4.1\end{tabular} \\
\begin{tabular}[c]{@{}c@{}}NVIDIA \\ GA100\end{tabular}   & \begin{tabular}[c]{@{}c@{}}A100 SXM4 \\ 80GB\end{tabular} & \begin{tabular}[c]{@{}c@{}}Intel(R) Xeon(R) \\ CPU @ 2.20GHz\\ 32 cores, 188 GB RAM\end{tabular} & Section 4.2   \\
\bottomrule
\end{tabular}}
\end{table}

\subsection{Parameters for \#MeToo Scaling Comparison}
\begin{table}[!ht]
\caption{Parameters used for TLDA in the \#MeToo scaling comparison}
\vspace{-0.8pt}
\label{tab:tlda_params}
\centering
\resizebox{1\linewidth}{!}{%
\begin{tabular}{ccccc}
\toprule
\textbf{\begin{tabular}[c]{@{}c@{}}Number of \\ Topics ($K$)\end{tabular}} & \textbf{Model} & \textbf{\begin{tabular}[c]{@{}c@{}}Topic Mixing \\ Parameter ($\alpha_0$)\end{tabular}} & \textbf{\begin{tabular}[c]{@{}c@{}}Learning \\ Rate ($\beta$)\end{tabular}} & \textbf{\begin{tabular}[c]{@{}c@{}}Whitening \\ Size ($D$) \end{tabular}} \\
\midrule
10                                                     & batched tLDA   & 0.001                                                                                   & 0.0005                                                                      & 40                                                                 \\
                                                                         & online tLDA    & 0.01                                                                                    & 0.0005                                                                      & 10                                                                 \\
20                                                     & batched tLDA   & 0.01                                                                                    & 0.00005                                                                     & 80                                                                 \\
                                                                         & online tLDA    & 0.001                                                                                   & 0.001                                                                       & 20                                                                 \\
40                                                      & batched tLDA   & 0.001                                                                                   & 0.001                                                                       & 160                                                                \\
                                                                         & online tLDA    & 0.001                                                                                   & 0.001                                                                       & 160                                                                \\
60                                                      & batched tLDA   & 0.01                                                                                    & 0.001                                                                       & 240                                                                \\
                                                                         & online tLDA    & 0.001                                                                                   & 0.001                                                                       & 240                                                                \\
80                                                      & batched tLDA   & 0.01                                                                                    & 0.00001                                                                     & 160                                                                \\
                                                                         & online tLDA    & 0.001                                                                                   & 0.0001                                                                      & 160                                                                \\
100                                                     & batched tLDA   & 0.001                                                                                   & 0.0005                                                                      & 100                                                                \\
                                                                         & online tLDA    & 0.001                                                                                   & 0.0005                                                                      & 100        \\
                                                                    \bottomrule
\end{tabular}
}
\end{table}

In this section, we record the hyperparameters for the TLDA and Gensim models used in the \#MeToo scaling comparison.

\begin{table}[!ht]
\caption{Non-default parameters used for gensim LDAMulticore in the \#MeToo scaling comparison}
\vspace{-0.8pt}
\label{tab:gensim_params}
\centering
\begin{tabular}{ccc}
\toprule
\textbf{Number of Topics} & \textbf{Passes} & \textbf{Iterations} \\
\midrule
10                        & 5               & 200                 \\
20                        & 5               & 200                 \\
40                        & 30              & 200                 \\
60                        & 50              & 200                 \\
80                        & 50              & 200                 \\
100                       & 50              & 200          \\
\bottomrule
\end{tabular}
\end{table}

In Table~\ref{tab:tlda_params}, we list the hyper-parameters used for TLDA in the \#MeToo scaling comparison (topic mixing parameter, learning rate, and whitening size). We also list, in Table~\ref{tab:gensim_params}, the non-default parameters we tuned for Gensim's LDA.

\subsection{Application to COVID-19 dataset}
In this section, we report an additional application to a large COVID-19 Data set

To further illustrate the scalability of the method, we analyze a second social media dataset of 260,752,483  pre-processed tweets (and 2,271,489,796 tokens) related to discussions around COVID-19 between January 1, 2020, and June 30, 2021. COVID-19 was a politically fraught event whose acute phase was experienced for months and years, affecting nearly every aspect of everyday life, society, and politics around the world. This period generated widespread discussion about public health policies,  education politics, presidential politics, and foreign policy. This dataset was generated as part of a long-term social media monitoring architecture that collects tweets in real time using certain keywords and hashtags related to COVID-19 \citep{DBLP:journals/corr/abs-2005-02442}.  These tweets in particular cover the discussion related to the politics of COVID-19, public health responses related to the pandemic, and COVID-19's cultural impact in 2020 and 2021. 

\subsubsection{Timing} 
Similarly to the \#MeToo analysis, we perform a grid-search over the number of topics K, topic mixing parameters $\alpha_0$, learning rates $\beta$, and  whitening sizes $D$. We use the online TLDA method because of the large scale of the data. However, due to the large size of the COVID dataset, computing the coherence is prohibitively time expensive, so we inspect the topics manually to find the following optimal parameters. We used $K$ = 5 topics; for the topic mixing parameter, we chose $\alpha_0$ = 0.0001. The learning rate was set to $\beta$ = $1\mathrm{e}{-5}$ and the whitening size to $20$.

In addition, to demonstrate that our method can scale to billions of documents, we use a large COVID dataset to simulate a dataset of $1,043,009,932$ documents. To do so, we loop through the entirety of the data 4 times and update the online version of the method on each batch once within each iteration.

\end{document}